\documentclass[letterpaper, 10 pt, conference]{ieeeconf}  

\usepackage{amsfonts}
\usepackage[margin=1.0in]{geometry}
\usepackage{mathtools}

\usepackage{placeins}
\usepackage{bbm}
\usepackage{algorithm}
\makeatletter
\newcommand{\thickhline}{%
    \noalign {\ifnum 0=`}\fi \hrule height 1pt
    \futurelet \reserved@a \@xhline
}
\allowdisplaybreaks
\usepackage{graphics}
\usepackage{bm}

\usepackage[hidelinks]{hyperref}
\usepackage[noend]{algpseudocode}
\usepackage{caption} 
\makeatletter
\usepackage{caption}
\usepackage{subcaption}
\usepackage{graphicx,xcolor} 
\usepackage[framemethod=tikz]{mdframed}
\usepackage{subfiles}
\usepackage{cite}
\usepackage{xcolor}
\usepackage{graphicx}
\usepackage{pifont}
\usepackage{xcolor}
\usepackage{graphicx}
\usepackage[utf8]{inputenc} 
\usepackage[T1]{fontenc}    
\usepackage{hyperref}       
\usepackage{url}            
\usepackage{booktabs}       
\usepackage{amsfonts}       
\usepackage{nicefrac}       
\usepackage{microtype}      
\usepackage[normalem]{ulem}
\usepackage{subfiles}

\IEEEoverridecommandlockouts                              

\usepackage{amsmath}
\usepackage{mathtools}

\title{\LARGE \bf
Bayesian Constraint Inference from User Demonstrations Based on Margin-Respecting Preference Models
}

\author{Dimitris Papadimitriou$^{1}$ and Daniel S. Brown$^{2}$
\thanks{$^{1}$Dimitris Papadimitriou is with the Department of Mechanical Engineering, 
        University of California at Berkeley, 
        {\tt\small dimitri@berkeley.edu}.}%
\thanks{$^{2}$Daniel S. Brown is with the School of Computing, University of Utah,
        {\tt\small daniel.s.brown@utah.edu}.}%
}

\begin{document}
\maketitle
\thispagestyle{empty}
\pagestyle{empty}

\begin{abstract}
It is crucial for robots to be aware of the presence of constraints in order to acquire safe policies. However, explicitly specifying all constraints in an environment can be a challenging task. State-of-the-art constraint inference algorithms learn constraints from demonstrations, but tend to be computationally expensive and prone to instability issues. In this paper, we propose a novel Bayesian method that infers constraints based on preferences over demonstrations. The main advantages of our proposed approach are that it \textit{1)} infers constraints without calculating a new policy at each iteration, \textit{2)} uses a simple and more realistic ranking of groups of demonstrations, without requiring pairwise comparisons over all demonstrations, and \textit{3)} adapts to cases where there are varying levels of constraint violation. Our empirical results demonstrate that our proposed Bayesian approach infers constraints of varying severity, more accurately than state-of-the-art constraint inference methods. Code and videos: \url{https://sites.google.com/berkeley.edu/pbicrl}.
\end{abstract}

\section{Introduction}\label{sec:intro}

Frequently, robots and other autonomous agents are required to act in environments under the presence of constraints. In such cases, policies can be obtained by using constrained reinforcement learning (RL) algorithms~\cite{achiam2017constrained,miryoosefi2022simple}. However, hand-specifying all constraints can be time consuming and error prone and some constraints may be user-dependent, precluding pre-specification of all relevant constraints. In environments in which some or all of the constraints are unknown, one might naturally aim at inferring them, a topic closely related to inverse reinforcement learning (IRL)~\cite{arora2021survey,abbeel2004apprenticeship,ng2000algorithms,wulfmeier2015maximum,ziebart2008maximum,lopes2009active,ramachandran2007bayesian,brown2019extrapolating, klein2012inverse}. 

Constraint inference in RL has been studied in both discrete~\cite{scobee2019maximum,papadimitrioubayesian} and continuous~\cite{malik2021inverse} state spaces. However, the majority of these methods suffer from high computational complexity. The reason for this is that these methods are iterative, and a new policy must be optimized at each iteration, incurring a significant computational cost. To alleviate this bottleneck, in this paper we formulate the constraint inference problem as a preference-based learning problem based on a Markov Chain Monte Carlo (MCMC) algorithm. More specifically, we utilize a preference-based likelihood function that can be evaluated efficiently, in order to calculate the likelihood of any proposed constraint. We further introduce a ranking over groups of preferences as a more natural framework for preferences in robotics applications. Finally, we wish to emphasize that not all constraints are of equal importance. For example, avoiding pedestrians vs. avoiding tree branches have very different levels of priority in autonomous driving. We borrow elements from maximum margin classification~\cite{liu2017sphereface,liu2016large,wang2018additive}  that enable our method to discriminate between constraints whose violations have varying degrees of consequences.


\section{Background and Related Work}\label{sec:rel_work}
Our work on constraint inference borrows elements from IRL and preference-based learning. 


\textbf{Constrained Reinforcement Learning}:\label{sec:cnstr_rl}
The typical constrained RL formulation obtains a policy $\pi$ by maximizing the expected reward while satisfying constraints in expectation.  Policies can be obtained by solving a constraint optimization problem~\cite{papadimitrioubayesian}.
One way to solve such an optimization problem, is by formulating the Lagrangian and solving the resulting min-max problem
\begin{align}
\min_{r_p {\leq} 0} \hspace{5pt} \max_\pi \hspace{5pt}
&\mathbb{E}_{a \sim \pi}\left[\sum_{t=1}^T\gamma^t r_{\boldsymbol{\theta}}(s,a)\right] \nonumber\\&+ 
r_p \left( \mathbb{E}_{a \sim \pi}\left[\sum_{t=1}^T \mathbb{I}_C(s,a)\right]\right),
\label{eq:lag_problem}
\end{align}
where $r_{\boldsymbol{\theta}}(s,a)$ is the reward function, parameterized by $\boldsymbol{\theta}$ and obtained at state $s$ for action $a$, $\mathbb{I}_C$ is the indicator function of constraint violation for state $s$ and action $a$ and $\gamma$ is the discount factor. The variable $r_p \in (-\infty, 0]$ denotes the Lagrange multiplier. This Lagrange multiplier can also be viewed as the average penalty that the agent incurs for violating constraints. We draw inspiration from this formulation to model the environment constraints. More specifically, we assume that constraints are ``soft" and that each time the agent violates one, the agent incurs a penalty reward $r_p$.

\textbf{Constraint Inference}:  The topic of constraint inference is a relatively new area of research in control and RL. From a control perspective, a common approach in literature is to infer constraint parameters by optimizing some metric based on the KKT conditions~\cite{chou2020learning,papadimitriou2023constraint}. In the RL community, maximum likelihood based methods have been developed that infer constraints in discrete environments~\cite{scobee2019maximum}. A Bayesian approach similar to ours, that learns the posterior distribution of constraints allowing for active learning methods, is developed in~\cite{papadimitrioubayesian}. The authors in~\cite{gaurav2022learning,malik2021inverse} propose alternating methods, of policy updates and constraint inference, that infer soft and hard constraints in continuous state-spaces. A maximum entropy approach for constraint inference is developed in~\cite{baert2023maximum} and an approach based on multi-task demonstrations in developed in~\cite{kim2023learning}. Finally, the authors in~\cite{lindner2023learning} construct safe sets after observing expert demonstrations in environments with varying underlying reward functions, while in~\cite{basich2023learning} the authors propose a constraint inference approach based only on sparse interventions from an expert. However, the previously mentioned approaches cannot infer constraints of varying magnitude, a capability that our approach provides.


\textbf{Preference-Based Learning}:
When demonstrations of varying degrees of optimality are available, it might be appropriate to utilize a preference-based model which allows the agent to infer reward parameters by just observing preferences over trajectories~\cite{wirth2017survey,lee2021b}. The authors in~\cite{akrour2011preference} propose a preference-based RL approach in which the agents utilize human feedback on their demonstrations in the form of ranking, to improve their policies. 
Based on the Bradley-Terry model~\cite{bradley1952rank}, many deep preference learning frameworks have been developed and studied~\cite{christiano2017deep,brown2019extrapolating,brown2020safe,lee2021pebble,bobu2023sirl,rafailov2024direct}. Preference learning has also been extended to multimodal reward functions~\cite{myers2022learning}, {offline RL~\cite{shin2023benchmarks}, and model-based RL~\cite{liu2023efficient}}. Generalizing from strict pairwise comparisons, the authors in~\cite{wilde2021learning} explore the benefits of providing feedback in the form of a scalar value, but do not consider constraint inference.

\section{Learning from Preferences}\label{sec:pref_learn}
This section presents the preference learning framework on which our model is based. We outline cases in which learning from preferences, even in the case when one has access to all possible pairwise comparisons, can be ineffective. We also introduce a less demanding group-wise comparison framework. Finally, we utilize tools from the large margin classification literature to train our models to mirror preferences of varying margins. 

Throughout this paper, we assume that we have access to a dataset $\mathcal{D}$ of $N$ demonstrations $\tau_i,\;i=1,\ldots N$, of varying quality. To simplify notation, we assume that the reward function and constraints, only depend on the state. We further assume that the reward function can be modeled as the sum of a known nominal reward and an unknown penalty reward that is associated with the constraints. Following prior work~\cite{papadimitrioubayesian, malik2021inverse}, we assume that the nominal reward is known and it is given by the inner product of the known nominal weights $\mathbf{w}_{n}\in\mathbb{R}^{N_{\phi}}$ and $N_{\phi}$ features $\boldsymbol{\phi}(s)$
\begin{align}
    r_{n}(s) = \mathbf{w}_{n}^{\top}\boldsymbol{\phi}(s),
\end{align}
and, similarly, the penalty reward is associated with the unknown penalty weights $\mathbf{w}_{p}\in\mathbb{R}^{N_{\phi}}$ as follows
\begin{align}
    r_{p}(s) = \mathbf{w}_{p}^{\top}\boldsymbol{\phi}(s).
\end{align}
We further associate each constraint with an unknown binary indicator variable $\mathbf{c}_p\in\mathbb{R}^{N_{\phi}}$ on top of the weight value $\mathbf{w}_p$. This binary indicator captures whether a feature is a constraint feature or not. The inspiration behind our constraint modeling approach with a binary variable, comes from integer programming techniques in constraint optimization problems where it is common to model non-convex constraints as a set of individual affine constraints~\cite{richards2005mixed,belotti2016handling}. 
We denote the cumulative environment reward as $r_{\boldsymbol{\theta}}(s)$, { with $\boldsymbol{\theta}=\{\mathbf{c}_p, \mathbf{w}_p\}$}, to designate the dependence on the unknown penalty weight $\mathbf{w}_p$ and unknown binary parameters $\mathbf{c}_p$. Thus, we have
\begin{align}
r_{\boldsymbol{\theta}}(s)=r_{n}(s)+r_{p}(s)=(\mathbf{w}_{n}+\mathbf{c}_p\circ\mathbf{w}_{p})^{\top}\boldsymbol{\phi}(s),
\end{align}
with $\circ$ denoting the Hadamard product.

Providing feedback in the form of pairwise comparisons is a natural way for humans to teach robots and other AI agents. The most frequently used model to express pairwise comparisons is the Bradley-Terry model~\cite{bradley1952rank}. The Bradley-Terry model assumes that the probability of demonstration $\tau_i$ being preferred to demonstration $\tau_j$ is modeled as
\begin{align}\label{eq:B-T_model}
    P(\tau_i\succ \tau_j)=\frac{e^{\beta\sum_{s\in\tau_i}r_{\boldsymbol{\theta}}(s)}}{e^{\beta\sum_{s\in\tau_i}r_{\boldsymbol{\theta}}(s)}+e^{\beta\sum_{s\in\tau_j}r_{\boldsymbol{\theta}}(s)}},
\end{align}
where $\beta$ denotes the inverse temperature parameter of the softmax function. The parameters $\boldsymbol{\theta}$ of the reward function are then learned in such a way so that the preferences among the demonstrations are satisfactorily preserved. Given that the demonstrations can be of varying length, we propose to consider the mean reward per demonstration, with $T_i$ denoting the length of demonstration $i$. Consequently, the log-likelihood of a certain $\mathbf{w}_{p}$ and $\mathbf{c}_p$ realization is given by
\begin{align}\label{eq:likeli_fn}
    &\mathcal{L}(\mathbf{w}_{p},\mathbf{c}_p)=\sum_{\tau_i\succ\tau_j} \log P(\tau_i\succ\tau_j)\nonumber\\&=\sum_{\tau_i\succ\tau_j}\log\frac{e^{\frac{\beta}{T_i}\sum_{s\in\tau_i}r_{\boldsymbol{\theta}}(s)}}{e^{\frac{\beta}{T_i}\sum_{s\in\tau_i}r_{\boldsymbol{\theta}}(s)}+e^{\frac{\beta}{T_j}\sum_{s\in\tau_j}r_{\boldsymbol{\theta}}(s)}}.
\end{align}
The log-likelihood objective function can then be used as  a metric to infer the unknown penalty weight vector $\mathbf{w}_{p}$ and binary variables $\mathbf{c}_p$ that best justify the user provided preferences. Although this formulation has been used extensively in the RL community~\cite{brown2019extrapolating,akrour2011preference,  christiano2017deep}, the Bradley-Terry choice model can also have its limitations.

\subsection{Limitations of Learning from Preferences}
\begin{figure}[h]
\centering
\begin{subfigure}[t]{0.24\textwidth}
                \centering   
                \includegraphics[trim={0 -0.9cm 0 0cm},clip,width=0.7\textwidth]{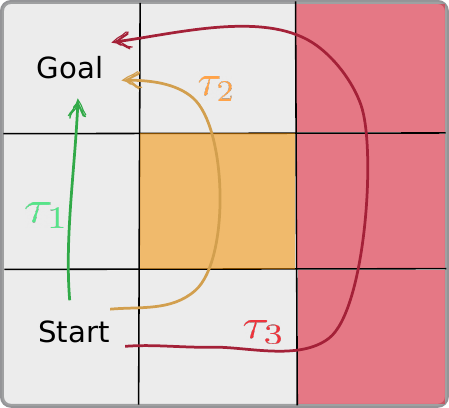}
                \caption{}                \label{fig:fig_example1}
        \end{subfigure}%
        \hspace{-0.5cm}
        \begin{subfigure}[t]{0.26\textwidth}
                \centering   
                \includegraphics[trim={-0.8 0cm 0 0cm},clip,width=0.86\textwidth]{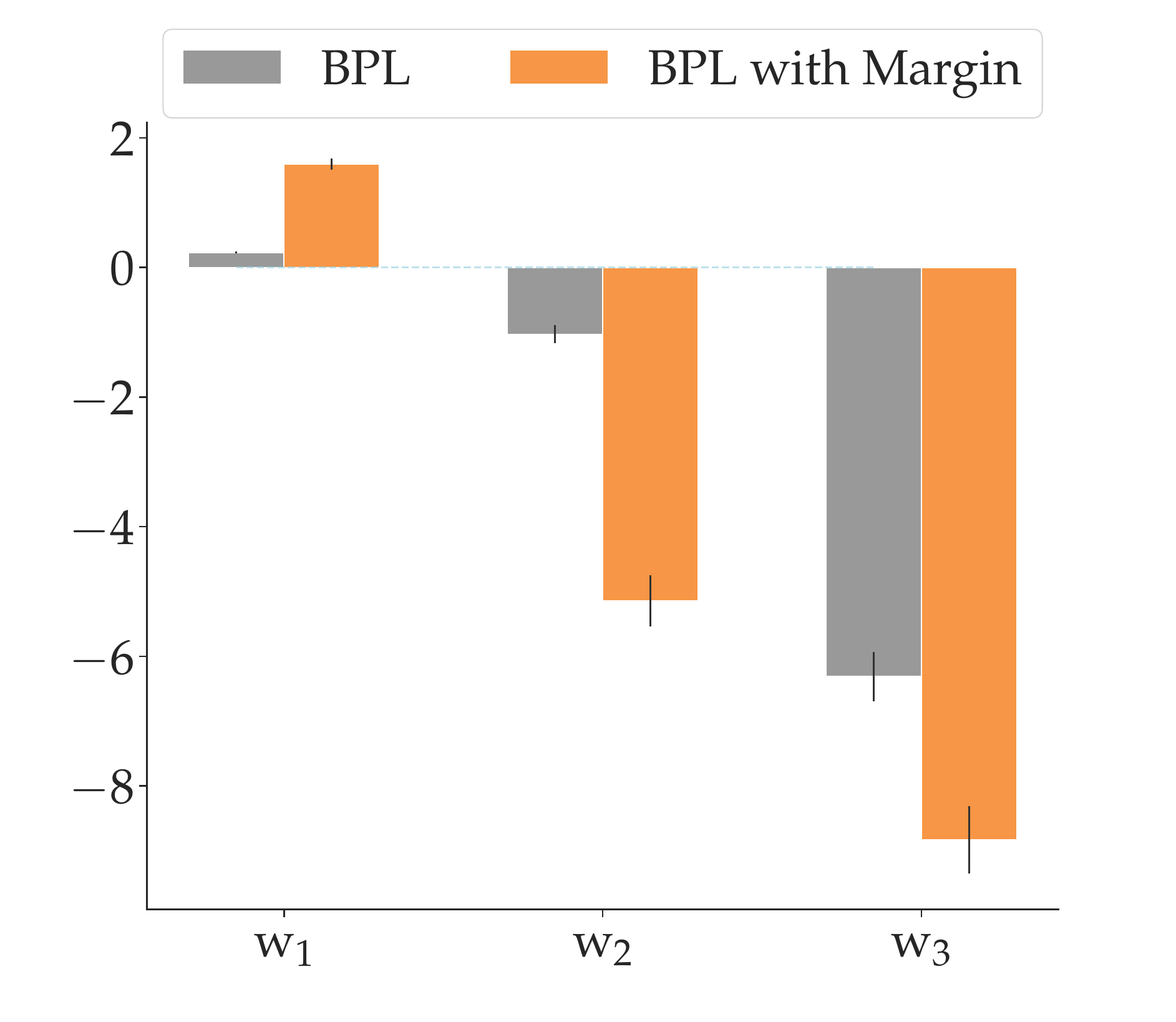}
                \caption{}
                \label{fig:bar_plot_ex}
         \end{subfigure}%
         \caption{{(\subref{fig:fig_example1})~Example of three different types of trajectories. A preferable (green), a bad (orange) and a slightly worse than orange (red). (\subref{fig:bar_plot_ex})~Weight values obtained from BPL and BPL with margins.}}
\label{fig:cont_human_prior}
\end{figure}

A drawback of the Bradley-Terry model is that it fails to capture ``how much more preferable" one choice is from another. To illustrate this, consider a simple preference learning task on the grid environment depicted in Figure~\ref{fig:fig_example1}. The agent must navigate from the start to the goal state in an environment in which the orange and the red states must be avoided, with the red considered slightly worse. The grey states can be considered good states. The expert provides the agent with the following complete list of pairwise comparisons, $\tau_1\succ\tau_2$, $\tau_2\succ\tau_3$ and $\tau_1\succ\tau_3$. 

For simplicity, we assume that each state on the $3\times3$ grid is associated with three binary features $(x,y,z)$. These are $(1,0,0)$  for the grey, $(0,1,0)$ for the orange and $(0,0,1)$ for the red states. We estimate $\mathbf{w}_{p}\in\mathbb{R}^3$ and $\mathbf{c}_p\in\mathbb{R}^3$ using Bayesian preference learning (BPL)~\cite{brown2020safe}, the details of which can be seen in Algorithm~\ref{alg:birl} 
in the Appendix. We run BPL $100$ independent times and report the average values of ${{w}_{pi}}, i=1,2,3$ in Figure~\ref{fig:bar_plot_ex} colored in grey. Conceptually, the Bradley-Terry model is a more appropriate modeling choice when the preferences among trajectories have similar significance. More specifically, the model fails to capture the fact that in our example, trajectory $\tau_1$ is significantly preferable to the others, i.e. $\tau_1\succ\succ\tau_2$ and $\tau_1\succ\succ\tau_3$. In this paper, we draw inspiration from the large margin classification literature~\cite{ montazery2017dominance,teso2016constructive,yuan2015non}, to propose a variation of the Bradley-Terry model that can be applied to preferences with non-uniform margins. 

\subsection{Margin-Respecting Preference Learning}
The likelihood function based on the Bradley-Terry model in Eq.~\eqref{eq:likeli_fn} treats every preference equally and does not enable us to enforce the relative strength of preferences or desired margins over predicted rewards. This limitation, showcased in Figure~\ref{fig:fig_example1}, is especially problematic in the constraint inference case when there are multiple constraints of varying priority or importance. 
These constraints can be hard to learn  even if there is a complete set of pairwise expert demonstration comparisons or a weighted likelihood function \cite{wirth2017survey}. 


In the context of traditional $K$-class CNN classification, where the last dense layer of a neural network is denoted with $\mathbf{W}$, the inputs in that layer are denoted with $\mathbf{f}_i, i=1,\ldots,N$ and the total number of data points is $N$, the softmax loss objective can be written as 
\begin{align}\label{eq:margin_likelihood}
\mathcal{L}(\mathbf{W})=
\frac{1}{N}\sum_{i=1}^N\log\frac{e^{\|\mathbf{W}_{y_i}\| \|\mathbf{f}_i\|cos(\theta_{{y_i,i}})}}{\sum_{j=1}^Ke^{\|\mathbf{W}_j\| \|\mathbf{f}_i\|cos({\theta_{j,i}})}},
\end{align}
where $\mathbf{W}_{y_i}$ denotes the row corresponding to the label $y_i$ {and $\theta_{y_i,i}$ is the angle between vectors $\mathbf{W}_{y_i}$ and $\mathbf{f}_i$}. Using this expression, the following objective function modification has been proposed in the literature~\cite{liu2017sphereface,liu2016large,wang2018additive}, that can encourage the creation of inter-class margins
\resizebox{.98\linewidth}{!}{
  \begin{minipage}{\linewidth}
\begin{align}\label{eq:margin_likelihood2}
&\mathcal{L}(\mathbf{W})=\nonumber \\&\frac{1}{N}\sum_{i=1}^N\log\frac{e^{\|\mathbf{W}_{y_i}\| \|\mathbf{f}_i\|\psi(\theta_{{y_i,i}})}}{e^{\|\mathbf{W}_{y_i}\|\|\mathbf{f}_i\|\psi({\theta_{y_i,i}})}+\sum_{j\neq y_i}^Ke^{\|\mathbf{W}_j\| \|\mathbf{f}_i\|cos({\theta_{j,i}})}},
\end{align}\end{minipage}}
with the use of an appropriate function $\psi(\cdot)$.
In the classification literature, the purpose of the $\psi(\cdot)$ function is to alter the angle between the rows of the last layer weight matrix so that the margin among classes is enlarged. In our work, we follow a similar approach to~\cite{wang2018additive} which assumes that $\psi({\theta_{y_i,i}})=cos({\theta_{y_i,i}})-m/(\|\mathbf{W}_{y_i}\|\|\mathbf{f}_i\|)$, where $m$ is a tunable hyperparameter, leading to a modified Bradley-Terry model that can be seen in~Eq.~\eqref{eq:likeli_fn_margin}. 

In order to tune $\psi(\cdot)$, some measure of how much more preferable a demonstration is to another is required. Our approach allows the demonstrator to specify the desired relationship of the margins between groups. For our example in Figure~\ref{fig:fig_example1}, the human specifies that the margin between $\tau_1$ and $\tau_2$ is approximately twice as large as that between $\tau_2$ and $\tau_3$. We then tune $\psi(\cdot)$ so that this relationship between the average rewards of the demonstrations is achieved under the learned constraint parameters. The improved results for an appropriate choice  of $m$ are shown in Figure~\ref{fig:bar_plot_ex} colored in orange. 

\subsection{Learning from Grouped Preferences}
Pairwise comparisons can be hard to obtain from humans for each possible pair of demonstrations. 
Hence, we propose to learn from coarsely grouped demonstrations as a more realistic alternative. In this case, a human can categorize demonstrations into distinct groups depending on their desirability. 
\begin{figure}[h]
    \centering
    \includegraphics[scale=0.36]{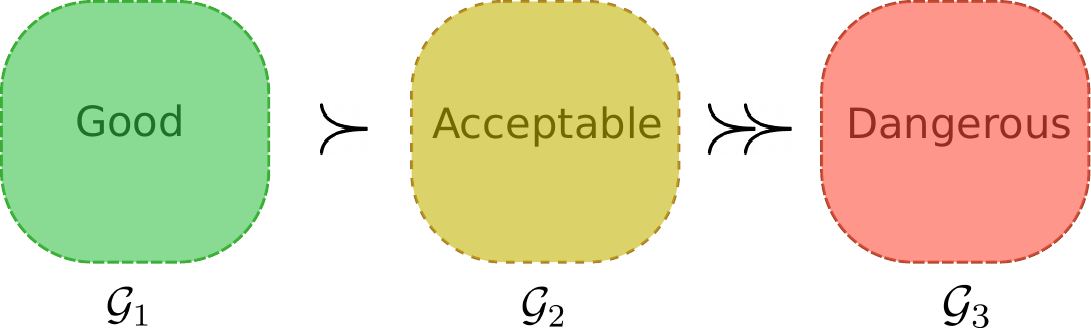}
    \caption{Illustrative example of grouped rankings of demonstrations with different inter-class margins.}    \label{fig:pref_illustration}
\end{figure}
Given rankings among groups $\mathcal{G}_i, i=1,\ldots,K$, where the groups are ordered from most to least preferred, the modified Bradley-Terry model and the likelihood function, for ${\tau_i\in \mathcal{G}_k,\tau_j\in\mathcal{G}_{\ell}}$ and $k<\ell$, can be written as
\begin{align}\label{eq:likeli_fn_margin}
     &{\mathcal{L}(\boldsymbol{\theta})}=\mathcal{L}(\mathbf{w}_{p},\mathbf{c}_p)=\hspace{-0.5cm}\sum\limits_{\substack{\forall \tau_i\in\mathcal{G}_k,\forall \tau_j\in\mathcal{G}_{\ell}\\\forall k<\ell}}\hspace{-0.5cm} \log P(\tau_i\succ\tau_j)\nonumber\\&=\hspace{-0.5cm}\sum\limits_{\substack{\forall \tau_i\in\mathcal{G}_k,\forall \tau_j\in\mathcal{G}_{\ell}\\\forall k<\ell}} \hspace{-0.5cm}\log\frac{e^{\frac{\beta}{T_i}\sum_{s\in\tau_i}r_{\boldsymbol{\theta}}(s)-m_{k\ell}}}{e^{\frac{\beta}{T_i}\sum_{s\in\tau_i}r_{\boldsymbol{\theta}}(s)-m_{k\ell}}+e^{\frac{\beta}{T_j}\sum_{s\in\tau_j}r_{\boldsymbol{\theta}}(s)}},
\end{align}
where $m_{k\ell}$ designates the margin parameter for groups $k$ and $\ell$. Note that the soft-max expression is evaluated between all $\tau_i\in\mathcal{G}_k$ and $\tau_j\in\mathcal{G}_{\ell}$. Thus, we leverage a small number of course-grained preference groupings over trajectories to automatically construct a large number of pairwise trajectory preferences. 

\section{Preference-Based Constraint Inference}\label{sec:pref_based_cnstr}
In this section, we outline the constraint inference approach for two cases, \textit{(1)} when the environment constraint features are known and \textit{(2)} when the constraint features assume a parametric form.

\subsection{Constraint Inference with Known Features}
Based on the likelihood in Eq.~\eqref{eq:likeli_fn_margin} we propose Preference-Based Bayesian Inverse Constraint Reinforcement Learning (PBICRL), an MCMC style algorithm that computes the posterior distribution of the constraints based on expert provided preferences over demonstrations. At each iteration, PBICRL randomly chooses a feature $j$ and then either samples a new binary constraint value or a new penalty weight value (using a Gaussian proposal distribution). The frequency with which a binary or a penalty weight value is sampled is determined by the sampling frequency $f_s$. After sampling, the new likelihood of the demonstrations is evaluated and compared with its value at the previous iteration to determine whether the sample is accepted or not. Upon completion, the algorithm returns the maximum a posteriori (MAP) penalty weight vector, as well as the MAP indicators of whether each feature is a constraint.
\begin{algorithm}[h]
\caption{PBICRL
}\label{alg:feat_inf}
\begin{algorithmic}[1]
\State \textbf{Parameters:} Number of iterations $k$, sampling frequency $f_s$, $\sigma$, margins $m_{k\ell}, \forall k\neq\ell\in \{1,\ldots, K\}$
\State Randomly sample: penalty weight $\mathbf{w}_{p}\in\mathbb{R}^{N_{\phi}}$, constraint feature $\mathbf{c}_{p}\in\{0,1\}^{N_{\phi}}$
\For{$i=1,\ldots,k$ }
\State Randomly sample feature $j$ from $\{1,\ldots,N_{\phi}\}$
\If{{$i$ mod $f_s$!=0}}
\State Set $\mathbf{c}_{p}'[j]=\lnot \mathbf{c}_{p}[j]$, $\mathbf{w}_{p}'=\mathbf{w}_{p}$
\Else
\State Set $\mathbf{w}_{p}'[j]= \mathbf{w}_{p}[j]+\mathcal{N}(0,\sigma)$, $\mathbf{c}_{p}'=\mathbf{c}_{p}$ 
\EndIf
\State Compute Likelihood using~(\ref{eq:likeli_fn_margin})
\If{$\log\mathcal{L}(\mathbf{w}_{p}',\mathbf{c}_{p}')\geq \log\mathcal{L}(\mathbf{w}_{p},\mathbf{c}_{p})$}
\State Set $\mathbf{w}_{p}=\mathbf{w}_{p}'$, $\mathbf{c}_{p}=\mathbf{c}_{p}'$ 
\Else
\State Set $\mathbf{w}_{p}=\mathbf{w}_{p}'$, $\mathbf{c}_{p}=\mathbf{c}_{p}'$ \\\;\;\;\;\;\;\;\;\;\;\;w.p. $\mathcal{L}(\mathbf{w}_{p}',\mathbf{c}_{p}')/\mathcal{L}(\mathbf{w}_{p},\mathbf{c}_{p})$
\EndIf
\EndFor
\State \textbf{Return} ${\mathbf{w}_{p}}, \mathbf{c}_{p}$
\end{algorithmic}
\end{algorithm}

\subsection{Constraint Inference with Unknown Features}
The previous section detailed the constraint inference approach in the case when the features were known.  In this section, we propose a variation of PBICRL that further allows for inferring the parametric form of the constraints. More specifically, we assume that the ${M_{\phi}}$ constraint features $\boldsymbol{\phi}^j_{{\vartheta}},\;j=1,\ldots,M_{\phi},$ now have functional forms parameterized by {$\boldsymbol{\vartheta}$}. In this case, we assume that the environment reward is given by
\begin{align}
r_{\boldsymbol{\theta}}(s)&=r_{n}(s)+r_{p,{\vartheta}}(s)\nonumber\\ &=\mathbf{w}_{n}^{\top}\boldsymbol{\phi}_n(s) + (\mathbf{c}_p\circ\mathbf{w}_{p})^{\top}\boldsymbol{\phi}_{{\vartheta}}(s),
\end{align}
with $\mathbf{w}_{p}\in\mathbb{R}^{M_{\phi}}$, $\mathbf{c}_p\in\mathbb{R}^{M_{\phi}}$ and {$\boldsymbol{\vartheta}\in\mathbb{R}^{M_{\vartheta}}$} being unknown parameters, {$\boldsymbol{\theta}=\{\mathbf{c}_p, \mathbf{w}_p, \boldsymbol{\vartheta}\}$,} and $\mathbf{w}_n\in\mathbb{R}^{N_{\phi}}$ still corresponding to the known nominal weight vector of the known features, which are now denoted with $\boldsymbol{\phi}_n$. This version of PBICRL, outlined in Algorithm~\ref{alg:feat_inf_unknown} in the Appendix, uses a modified sampling scheme in which the unknown constraint parameters $\boldsymbol{\vartheta}$ are also sampled and included in the calculation of the likelihood in Eq.~(\ref{eq:likeli_fn_margin}).

\section{Experiments}\label{sec:sims}
We utilize four different environments to quantify the performance of PBICRL: \textit{(1)} a 2D point mass (PM) navigation environment, \textit{(2)} the Fetch-Reach (FR) environment from OpenAI Gym and \textit{(3)} the HalfCheetah (HC) and \textit{(4)} the Ant environment provided in Gymnasium~\cite{gymnasium_robotics2023github}. Demonstrations in all environments were gathered from policies obtained with the Soft Actor-Critic (SAC) algorithm~\cite{haarnoja2018soft}.
\begin{figure}[t]
\centering
        \begin{subfigure}[t]{0.25\textwidth}
                \centering   
                \includegraphics[trim={0.0cm 0.0cm 0 0cm},clip,width=1.07\textwidth]{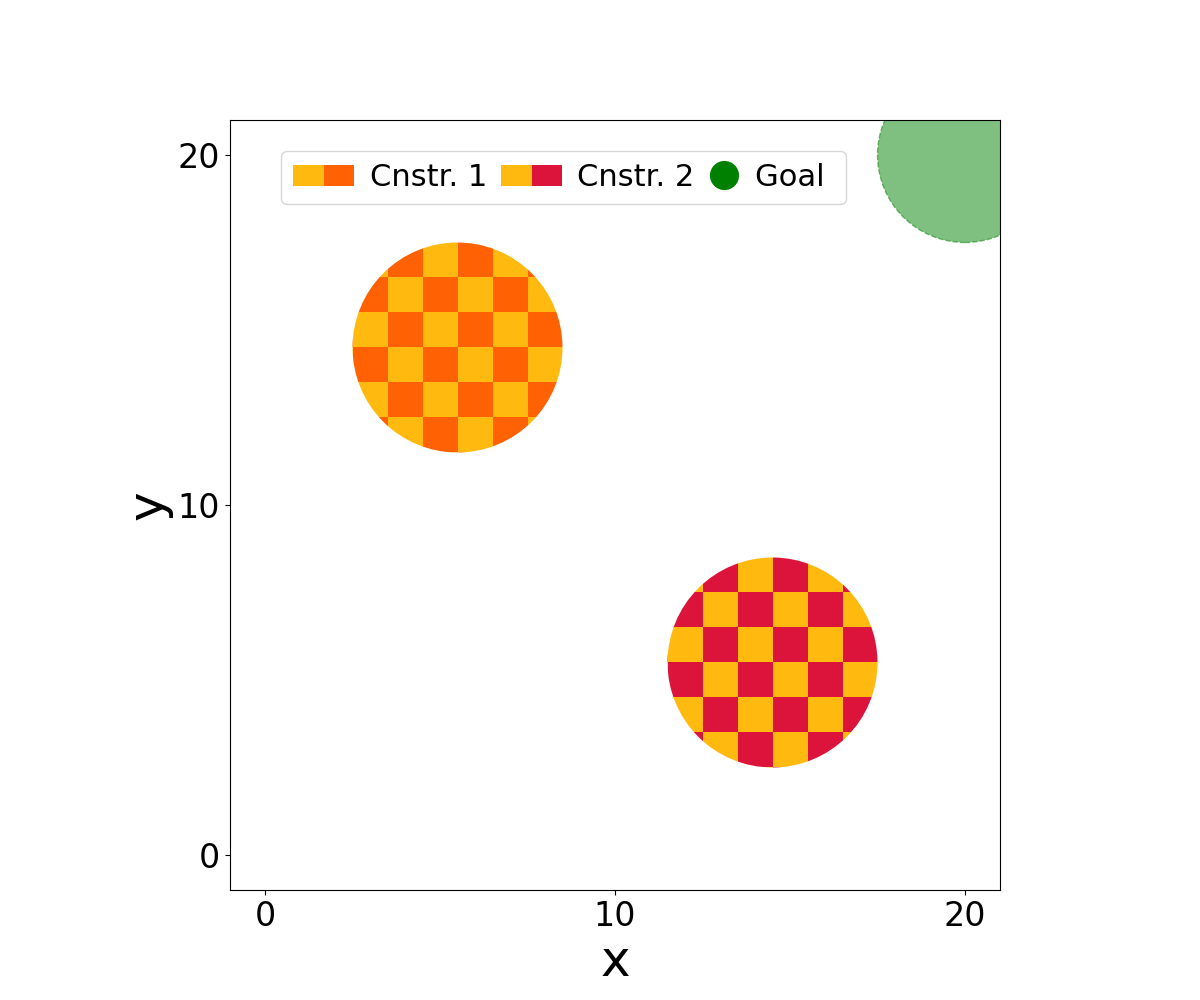}
                \caption{}    \label{fig:2d_env_2}
        \end{subfigure}%
        \begin{subfigure}[t]{0.244\textwidth}
                \centering    \includegraphics[trim={0cm -0.85cm 0.0 0.0cm},clip,width=0.75\textwidth]{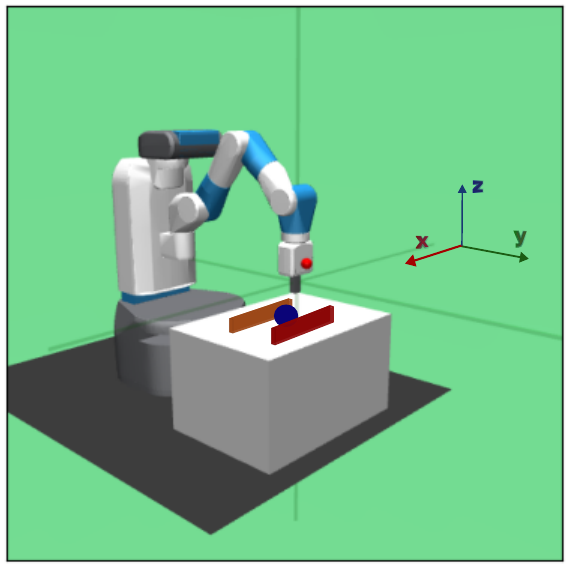}
                \caption{}    \label{fig:fetch_env_2}
        \end{subfigure}
        \\
                
         \caption{2D point mass navigational environment (\subref{fig:2d_env_2}) and Fetch-Reach robot (\subref{fig:fetch_env_2}) environments.}
\label{fig:point_mass_comp_100}
\end{figure}
\subsection{Point Mass Environment}
First, we study a two-dimensional environment, pictured in Figure~\ref{fig:2d_env_2}, with the agent following a point mass model that navigates from some starting state towards the goal set, at the top right corner, while avoiding certain obstacles. At each time step the agent accrues a reward proportional to the inverse of its distance from the center of the goal set while at the same time accruing a living cost of $-1$. Once the point mass enters the goal region the episode is completed. There are a total of two constraints, the orange and the red one. The agent incurs a penalty of $r_{p1}$ and $r_{p2}$ upon entering the orange and red constraint regions, respectively. Furthermore, the actual unknown penalties satisfy $r_{p2}\ll r_{p1}$, as demonstrations that violate the red constraint are considered highly suboptimal. {Although the penalties are unknown, we aim at recovering them accurately based on the expert provided pairwise comparisons and the relative margins among them.} There are four features in the environment, namely: \textit{(1)} the inverse distance from the target center, \textit{(2)} an indicator variable on whether the orange constraint is violated,
\textit{(3)} an indicator variable on whether the red constraint is violated and
\textit{(4)} a binary variable indicating whether the target has not been reached yet. The ground-truth weight vector associated with the features is $\mathbf{w}_{}=\mathbf{w}_n+\mathbf{w}_p=[1,-10,-100,-1]$. 

Similar to~\cite{papadimitrioubayesian,scobee2019maximum}, we assume that we only have access to the nominal weight vector $\mathbf{w}_{n}=[1,0,0,-1]$. Using PBICRL we infer $\mathbf{w}_{p}$ and $\mathbf{c}_p$.
\begin{figure}[t]
\centering
\begin{subfigure}[t]{0.25\textwidth}
                \centering   
                \hspace{-0.6cm}\includegraphics[width=1.08\textwidth]{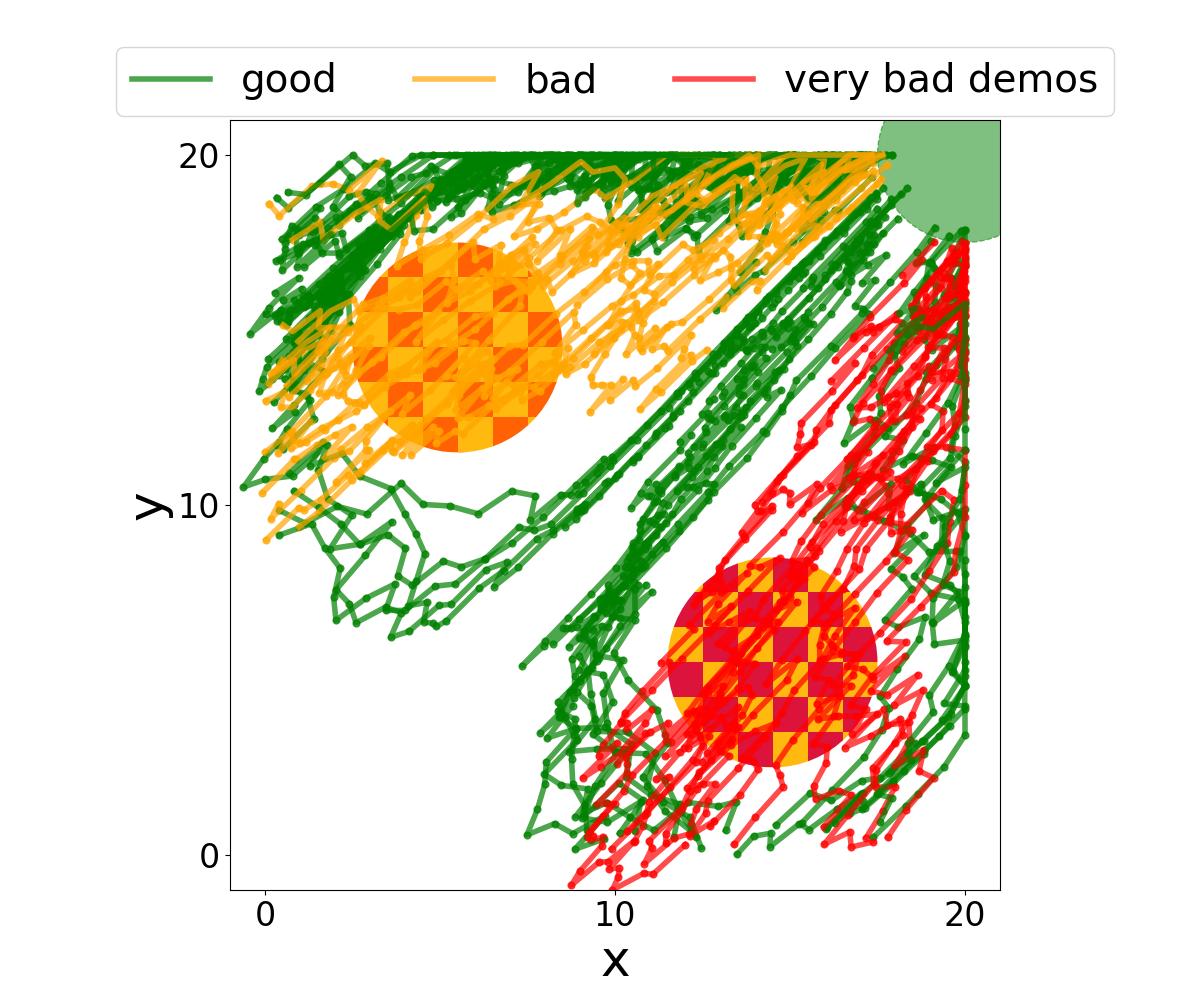}
                \caption{}                \label{fig:2d_traj}
        \end{subfigure}%
        \begin{subfigure}[t]{0.26\textwidth}
                \centering   
                \includegraphics[width=1\textwidth]{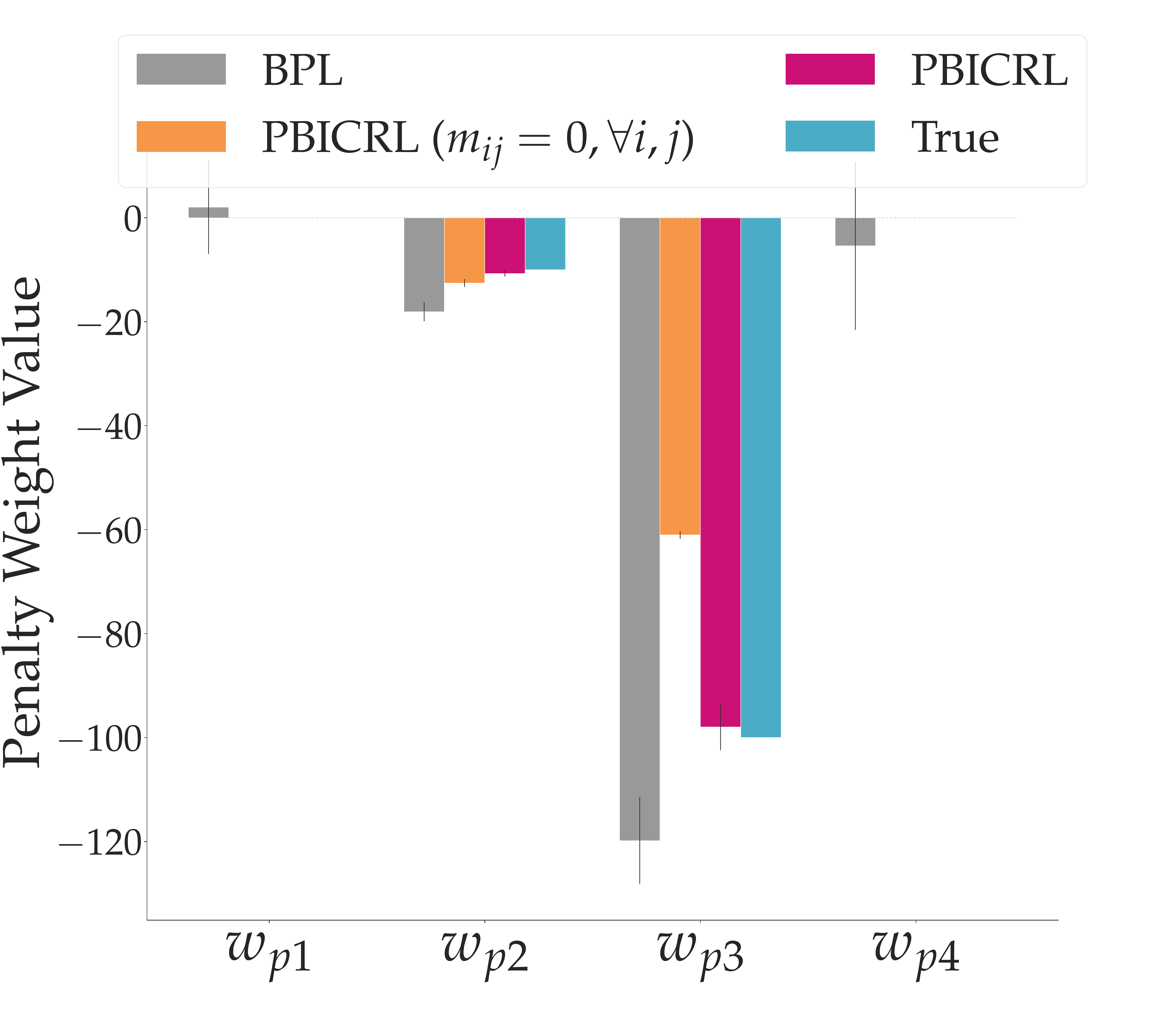}
                \caption{}    \label{fig:2d_results}
         \end{subfigure}      \caption{(\subref{fig:2d_traj})~2D point mass environment demonstrations. (\subref{fig:2d_results}) Inference results for point mass environment. Results averaged over $5$ seeds.}
\label{fig:cont_human_prior}
\end{figure}
Figure~\ref{fig:2d_traj} depicts some demonstrations that are categorized in three groups $\mathcal{G}_1$, $\mathcal{G}_2$ and $\mathcal{G}_3$, representing good, bad and very bad sets of demonstrations, respectively. 
We compare the performance of PBICRL with BPL~\cite{brown2020safe}, the details of which can be seen in Algorithm~\ref{alg:birl} in the Appendix. In BPL we also assume knowledge of the nominal reward weight vector $\mathbf{w}_{n}$ and the algorithm infers the penalty weight vector $\mathbf{w}_{p}$. The main difference between PBICRL and BPL is that the latter is missing the binary variables that signify the presence or absence of constraints.

We first run PBICRL, with all the margins $m_{ij},\forall i,j$ set to zero, and BPL using $60$ demonstrations for each of the three groups and we average the results over $5$ seeds. Figure~\ref{fig:2d_results} shows the inferred weights for PBICRL and BPL, respectively, along with the true values. Bayesian preference learning (BPL) fails to infer the correct weights, as for instance $w_{p1}$ and $w_{p4}$ should both be zero. On the other hand, the performance of PBICRL is significantly better, as it correctly identifies the two constrained features and the corresponding weight values are close to the true ones. 

To capture how much undesirable the red constraint is, which should translate to a lower $w_{p3}$ value, we further need to tune the $m_{ij}$ margin parameters. To tune these margins, our method can utilize additional  feedback from the expert. In this example, the expert provides feedback that the relative margins among demonstrations in $\mathcal{G}_2$ and $\mathcal{G}_3$ are approximately four times the magnitude of those between $\mathcal{G}_1$ and $\mathcal{G}_2$. The parameters $m_{ij}$ are automatically tuned so the cumulative rewards of the demonstrations under the inferred parameters approximately satisfy this condition. After tuning, the new inferred weights, as seen in Figure~\ref{fig:2d_results}, improve significantly.

\subsection{Fetch-Reach Robot}
In this section, we utilize the Fetch-Reach robotic environment~\cite{gymnasium_robotics2023github} and train it with a SAC policy to navigate towards the blue goal set, as shown in Figure~\ref{fig:fetch_env_2}. The environment contains two rectangular constraint regions: the orange one is associated with a ground-truth violation penalty of $-20$ and the considerably worse red one has a violation penalty of $-100$. The blue colored sphere designates the goal set towards which the robot should navigate. The environment has four features, namely: \textit{(1)} the inverse distance from the center of the goal set,
\textit{(2)} a binary indicator for violating the orange constraint, \textit{(3)} a binary indicator for violating the red constraint, and \textit{(4)} a binary living indicator. The original ground-truth weight vector $\mathbf{w}\in\mathbb{R}^4$ is $[0.1,-20,-100,-5]$, while the known nominal weight vector is $\mathbf{w}_n=[0.1,0,0,-5]$ and the unknown constraint violation parameters $\mathbf{w}_p$ and $\mathbf{c}_p$ are to be inferred.

\begin{figure}[t]
\centering        
        \begin{subfigure}[t]{0.25\textwidth}
                \centering   
                \includegraphics[trim={3.5cm 1cm 0 1cm},clip,scale=0.4]{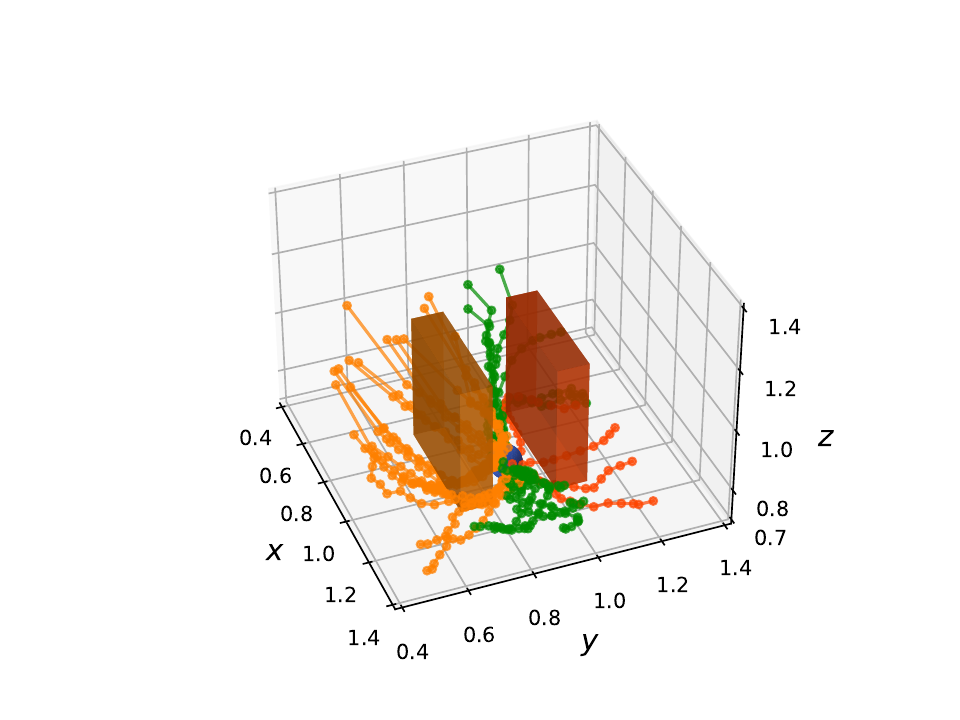}
                \caption{}
                \label{fig:fetch_traj}
         \end{subfigure}%
         \begin{subfigure}[t]{0.25\textwidth}
                \centering   
            \includegraphics[trim={0cm 0cm 0 1cm},clip,width=0.99\textwidth]{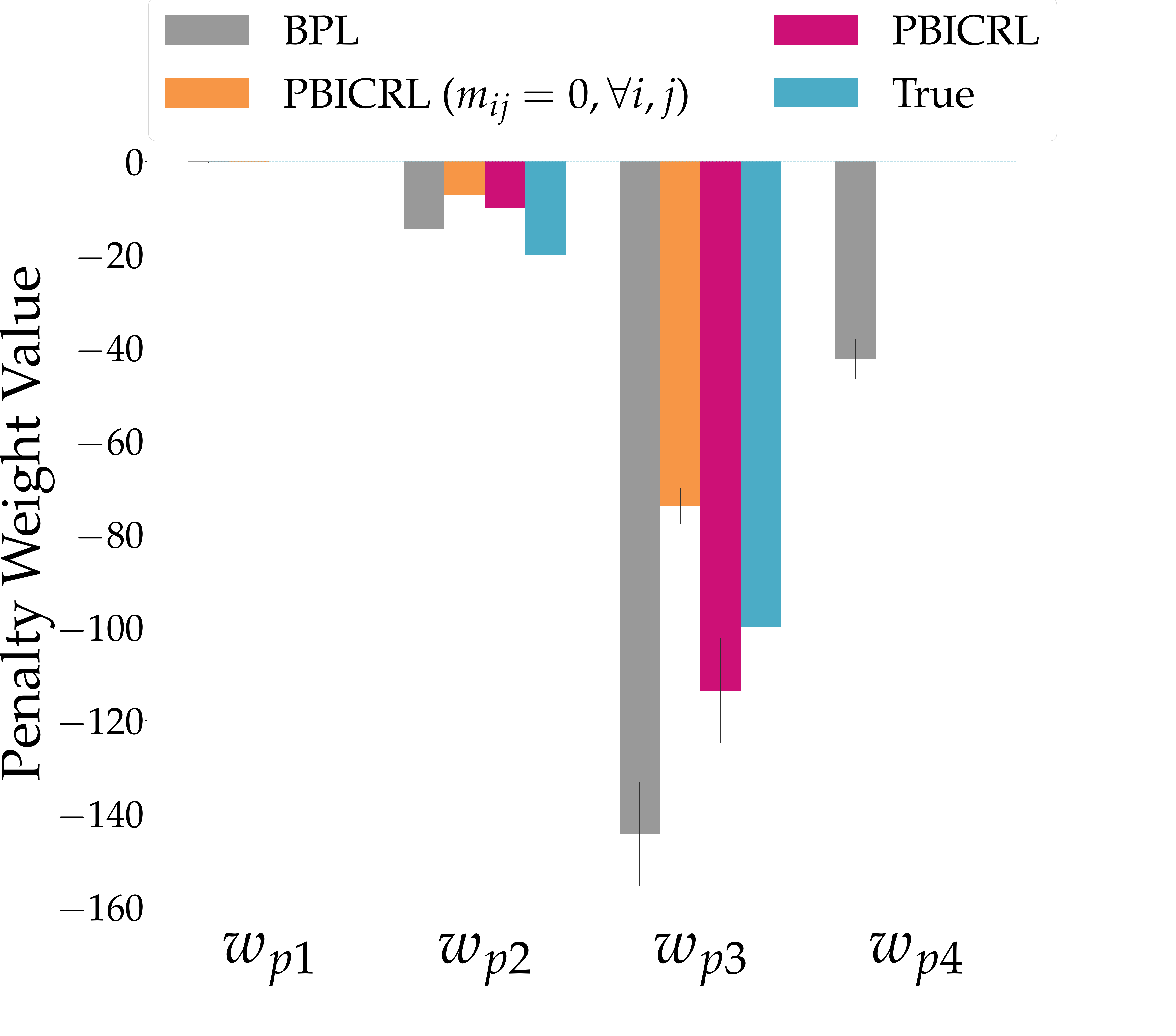}
                \caption{}                \label{fig:fetch_res}
        \end{subfigure}%
         \caption{(\subref{fig:fetch_traj}) Fetch-Reach robot demonstrations. (\subref{fig:fetch_res}) Inference results for Fetch-Reach environment. Results averaged over $5$ seeds.}
\label{fig:cont_human_prior}
\end{figure}

Figure~\ref{fig:fetch_res} shows the results for the inferred constraint weights. The $m_{ij}$ parameters were tuned under the assumption that the user provides feedback that the relative margins between the two pairs of groups $\mathcal{G}_1, \mathcal{G}_2$ and $\mathcal{G}_2, \mathcal{G}_3$ are approximately the same. The original average rewards of the demonstrations, along with the ones for the inferred weights under $m_{ij}=0, \forall i,j$ and after tuning $m_{ij}$, can be seen in the Appendix for both the 2D and Fetch-Reach environments.
\subsection{HalfCheetah and Ant with Parametric Constraints}
In this section we consider parametric 1D halfspace constraint functions of the form $z\leq \vartheta$, where PBICRL, must also infer the parameter $\vartheta$. We focus on the $z$ dimension as this is the primary axis along which the agent is incentivized to run forwards. This framework further allows us to compare our method to state-of-the-art constraint inference techniques such as~\cite{gaurav2022learning,malik2021inverse}, that also infer a single 1D constraint with a parametric form. Given that in our experiments constraint violation is technically possible, we focus on comparing with~\cite{gaurav2022learning}, which is a soft constraint inference algorithm.

In the HalfCheetah and Ant environments~\cite{todorov2012mujoco} we test the performance of PBICRL in a scenario with a single constraint function, similar to~\cite{gaurav2022learning}. More specifically, we assume that there exists an unobserved constraint at a particular location $z$.  For HalfCheetah, we assume that the agent faces a constraint at the $z=8$ location and whenever $z\geq 8$ the agent incurs a penalty reward of $-50$. For the Ant environment, we assume that the states for which $z\geq 10$ are associated with an unobserved constraint reward of $-100$. 

The known features $\boldsymbol{\phi}_n$ of the HalfCheetah and Ant environments are the z-coordinate difference between two time steps and the square norm of the action vector. The corresponding nominal weights are $\mathbf{w}_n=[20,-0.1]$. In this setting, PBICRL infers the unknown constraint location $\vartheta$ as well as the $\mathbf{c}_p$, $\mathbf{w}_p$ parameters associated with it.
We compare the inferred constraints of PBICRL with Inverse Constraint Learning (ICL)~\cite{gaurav2022learning} by reporting the Constraint Mean Squared Error (CMSE), which quantifies the error between the location of the actual and the inferred constraints. For PBICRL, $100$ good and $100$ bad demonstrations were provided. In the ICL method, we used $200$ good demonstrations some of which violated the constraint. Table~\ref{table:cmse} shows the CMSE values for both methods. PBICRL consistently estimates a constraint location very close to the true one, while ICL tends to infer a constraint close to the starting states of the demonstrations.  
\begin{table}[H]
\caption{CMSE for PBICRL and ICL.}
\centering
\begin{tabular}{cccccccc}
\thickhline
 Env. & & \multicolumn{3}{c}{ICL} & \multicolumn{3}{c}{PBICRL}\\ \cmidrule(lr){1-2} \cmidrule(lr){3-5} \cmidrule(lr){6-8}
 HC & & & $69.74\pm 0.40$   &  &  & $\bm{0.073}\pm\bm{0.12}$  &  \\ 
  Ant & & & $103.37\pm 0.21$   &  &  &  \bm{$0.20$} $\pm$ \bm{$0.29$}  &  \\ 
 \thickhline
\end{tabular}
\label{table:cmse}
\end{table}
Another benefit of our approach lies in its time complexity. Prior work alternates between policy learning and constraint inference~\cite{gaurav2022learning,malik2021inverse} leading to high time complexities. By contrast, our approach infers constraints in the above environments using an order of magnitude less time. Table~\ref{table:run_times} shows the time complexity of ICL and PBICRL for the HalfCheetah and Ant environments. All results have been averaged over $5$ seeds. 
\begin{table}[H]
\caption{Runtime in seconds for PBICRL and ICL.}
\centering
\begin{tabular}{cccccccc}
\thickhline
 Env. & & \multicolumn{3}{c}{ICL} & \multicolumn{3}{c}{PBICRL}\\ \cmidrule(lr){1-2} \cmidrule(lr){3-5} \cmidrule(lr){6-8}
 HC & & & $20650\pm 2982$   &  &  & $\bm{3298}\pm \bm{659}$ &  \\ 
  Ant & & & $22703$ $\pm$ {$2961$}   &  &  & \bm{$3165$} $\pm$ \bm{$12$} &  \\ 
 \thickhline
\end{tabular}
\label{table:run_times}
\end{table}

\section{Conclusion and Future Directions}\label{sec:conc}
We presented a novel Bayesian constraint learning algorithm (PBICRL) that utilizes preferences over groups of demonstrations. 
We also extended the classic Bradley-Terry choice model to allow different margins between preferences. 
Our empirical results show that PBICRL can more accurately and more efficiently infer constraints in environments under both known and unknown features than prior state-of-the-art approaches for both constraint learning and Bayesian preference learning. 

One of the benefits of our Bayesian inference approach is that the posterior distribution can provide information about uncertainty and hence, could be used to design active learning approaches as well as design policies that satisfy certain safety criteria. One particular direction we consider interesting is the utilization of the preference margins, which inherently quantify the severity of individual constraints, in an active learning scheme that uses a query function that takes into consideration those margins. Future work also includes extending our approach to higher dimensional environments using unsupervised learning~\cite{chen2020simple} and human-guided representation alignment~\cite{brown2020safe,bobu2023sirl,mattson2023leveraging} to learn feature representations that can be used for constraint learning.

\bibliographystyle{IEEEtran}
\bibliography{IEEEabrv,references}

\begin{thebibliography}{10}
\providecommand{\url}[1]{#1}
\csname url@rmstyle\endcsname
\providecommand{\newblock}{\relax}
\providecommand{\bibinfo}[2]{#2}
\providecommand\BIBentrySTDinterwordspacing{\spaceskip=0pt\relax}
\providecommand\BIBentryALTinterwordstretchfactor{4}
\providecommand\BIBentryALTinterwordspacing{\spaceskip=\fontdimen2\font plus
\BIBentryALTinterwordstretchfactor\fontdimen3\font minus \fontdimen4\font\relax}
\providecommand\BIBforeignlanguage[2]{{%
\expandafter\ifx\csname l@#1\endcsname\relax
\typeout{** WARNING: IEEEtran.bst: No hyphenation pattern has been}%
\typeout{** loaded for the language `#1'. Using the pattern for}%
\typeout{** the default language instead.}%
\else
\language=\csname l@#1\endcsname
\fi
#2}}

\bibitem{achiam2017constrained}
J.~Achiam, D.~Held, A.~Tamar, and P.~Abbeel, ``Constrained policy optimization,'' in \emph{International conference on machine learning}.\hskip 1em plus 0.5em minus 0.4em\relax PMLR, 2017, pp. 22--31.

\bibitem{miryoosefi2022simple}
S.~Miryoosefi and C.~Jin, ``A simple reward-free approach to constrained reinforcement learning,'' in \emph{International Conference on Machine Learning}.\hskip 1em plus 0.5em minus 0.4em\relax PMLR, 2022, pp. 15\,666--15\,698.

\bibitem{arora2021survey}
S.~Arora and P.~Doshi, ``A survey of inverse reinforcement learning: Challenges, methods and progress,'' \emph{Artificial Intelligence}, vol. 297, p. 103500, 2021.

\bibitem{abbeel2004apprenticeship}
P.~Abbeel and A.~Y. Ng, ``Apprenticeship learning via inverse reinforcement learning,'' in \emph{Proceedings of the twenty-first international conference on Machine learning}, 2004, p.~1.

\bibitem{ng2000algorithms}
A.~Y. Ng, S.~Russell, \emph{et~al.}, ``Algorithms for inverse reinforcement learning.'' in \emph{Icml}, vol.~1, 2000, p.~2.

\bibitem{wulfmeier2015maximum}
M.~Wulfmeier, P.~Ondruska, and I.~Posner, ``Maximum entropy deep inverse reinforcement learning,'' \emph{arXiv preprint arXiv:1507.04888}, 2015.

\bibitem{ziebart2008maximum}
B.~D. Ziebart, A.~L. Maas, J.~A. Bagnell, A.~K. Dey, \emph{et~al.}, ``Maximum entropy inverse reinforcement learning.'' in \emph{Aaai}, vol.~8.\hskip 1em plus 0.5em minus 0.4em\relax Chicago, IL, USA, 2008, pp. 1433--1438.

\bibitem{lopes2009active}
M.~Lopes, F.~Melo, and L.~Montesano, ``Active learning for reward estimation in inverse reinforcement learning,'' in \emph{Machine Learning and Knowledge Discovery in Databases: European Conference, ECML PKDD 2009, Bled, Slovenia, September 7-11, 2009, Proceedings, Part II 20}.\hskip 1em plus 0.5em minus 0.4em\relax Springer Berlin Heidelberg, 2009, pp. 31--46.

\bibitem{ramachandran2007bayesian}
D.~Ramachandran and E.~Amir, ``Bayesian inverse reinforcement learning.'' in \emph{IJCAI}, vol.~7, 2007, pp. 2586--2591.

\bibitem{brown2019extrapolating}
D.~Brown, W.~Goo, P.~Nagarajan, and S.~Niekum, ``Extrapolating beyond suboptimal demonstrations via inverse reinforcement learning from observations,'' in \emph{International conference on machine learning}.\hskip 1em plus 0.5em minus 0.4em\relax PMLR, 2019, pp. 783--792.

\bibitem{klein2012inverse}
E.~Klein, M.~Geist, B.~Piot, and O.~Pietquin, ``Inverse reinforcement learning through structured classification,'' \emph{Advances in neural information processing systems}, vol.~25, 2012.

\bibitem{scobee2019maximum}
D.~R. Scobee and S.~S. Sastry, ``Maximum likelihood constraint inference for inverse reinforcement learning,'' \emph{arXiv preprint arXiv:1909.05477}, 2019.

\bibitem{papadimitrioubayesian}
D.~Papadimitriou, U.~Anwar, and D.~S. Brown, ``Bayesian methods for constraint inference in reinforcement learning,'' \emph{Transactions on Machine Learning Research}.

\bibitem{malik2021inverse}
S.~Malik, U.~Anwar, A.~Aghasi, and A.~Ahmed, ``Inverse constrained reinforcement learning,'' in \emph{International Conference on Machine Learning}.\hskip 1em plus 0.5em minus 0.4em\relax PMLR, 2021, pp. 7390--7399.

\bibitem{liu2017sphereface}
W.~Liu, Y.~Wen, Z.~Yu, M.~Li, B.~Raj, and L.~Song, ``Sphereface: Deep hypersphere embedding for face recognition,'' in \emph{Proceedings of the IEEE conference on computer vision and pattern recognition}, 2017, pp. 212--220.

\bibitem{liu2016large}
W.~Liu, Y.~Wen, Z.~Yu, and M.~Yang, ``Large-margin softmax loss for convolutional neural networks,'' \emph{arXiv preprint arXiv:1612.02295}, 2016.

\bibitem{wang2018additive}
F.~Wang, J.~Cheng, W.~Liu, and H.~Liu, ``Additive margin softmax for face verification,'' \emph{IEEE Signal Processing Letters}, vol.~25, no.~7, pp. 926--930, 2018.

\bibitem{chou2020learning}
G.~Chou, N.~Ozay, and D.~Berenson, ``Learning constraints from locally-optimal demonstrations under cost function uncertainty,'' \emph{IEEE Robotics and Automation Letters}, vol.~5, no.~2, pp. 3682--3690, 2020.

\bibitem{papadimitriou2023constraint}
D.~Papadimitriou and J.~Li, ``Constraint inference in control tasks from expert demonstrations via inverse optimization,'' \emph{arXiv preprint arXiv:2304.03367}, 2023.

\bibitem{gaurav2022learning}
A.~Gaurav, K.~Rezaee, G.~Liu, and P.~Poupart, ``Learning soft constraints from constrained expert demonstrations,'' \emph{arXiv preprint arXiv:2206.01311}, 2022.

\bibitem{baert2023maximum}
M.~Baert, P.~Mazzaglia, S.~Leroux, and P.~Simoens, ``Maximum causal entropy inverse constrained reinforcement learning,'' \emph{arXiv preprint arXiv:2305.02857}, 2023.

\bibitem{kim2023learning}
K.~Kim, G.~Swamy, Z.~Liu, D.~Zhao, S.~Choudhury, and Z.~S. Wu, ``Learning shared safety constraints from multi-task demonstrations,'' \emph{arXiv preprint arXiv:2309.00711}, 2023.

\bibitem{lindner2023learning}
D.~Lindner, X.~Chen, S.~Tschiatschek, K.~Hofmann, and A.~Krause, ``Learning safety constraints from demonstrations with unknown rewards,'' \emph{arXiv preprint arXiv:2305.16147}, 2023.

\bibitem{basich2023learning}
C.~Basich, S.~Mahmud, and S.~Zilberstein, ``Learning constraints on autonomous behavior from proactive feedback,'' in \emph{2023 IEEE/RSJ International Conference on Intelligent Robots and Systems (IROS)}.\hskip 1em plus 0.5em minus 0.4em\relax IEEE, 2023, pp. 3680--3687.

\bibitem{wirth2017survey}
C.~Wirth, R.~Akrour, G.~Neumann, J.~F{\"u}rnkranz, \emph{et~al.}, ``A survey of preference-based reinforcement learning methods,'' \emph{Journal of Machine Learning Research}, vol.~18, no. 136, pp. 1--46, 2017.

\bibitem{lee2021b}
K.~Lee, L.~Smith, A.~Dragan, and P.~Abbeel, ``B-pref: Benchmarking preference-based reinforcement learning,'' in \emph{Thirty-fifth Conference on Neural Information Processing Systems Datasets and Benchmarks Track (Round 1)}, 2021.

\bibitem{akrour2011preference}
R.~Akrour, M.~Schoenauer, and M.~Sebag, ``Preference-based policy learning,'' in \emph{Machine Learning and Knowledge Discovery in Databases: European Conference, ECML PKDD 2011, Athens, Greece, September 5-9, 2011. Proceedings, Part I 11}.\hskip 1em plus 0.5em minus 0.4em\relax Springer, 2011, pp. 12--27.

\bibitem{bradley1952rank}
R.~A. Bradley and M.~E. Terry, ``Rank analysis of incomplete block designs: I. the method of paired comparisons,'' \emph{Biometrika}, vol.~39, no. 3/4, pp. 324--345, 1952.

\bibitem{christiano2017deep}
P.~F. Christiano, J.~Leike, T.~Brown, M.~Martic, S.~Legg, and D.~Amodei, ``Deep reinforcement learning from human preferences,'' \emph{Advances in neural information processing systems}, vol.~30, 2017.

\bibitem{brown2020safe}
D.~Brown, R.~Coleman, R.~Srinivasan, and S.~Niekum, ``Safe imitation learning via fast bayesian reward inference from preferences,'' in \emph{International Conference on Machine Learning}.\hskip 1em plus 0.5em minus 0.4em\relax PMLR, 2020, pp. 1165--1177.

\bibitem{lee2021pebble}
K.~Lee, L.~Smith, and P.~Abbeel, ``Pebble: Feedback-efficient interactive reinforcement learning via relabeling experience and unsupervised pre-training,'' \emph{arXiv preprint arXiv:2106.05091}, 2021.

\bibitem{bobu2023sirl}
A.~Bobu, Y.~Liu, R.~Shah, D.~S. Brown, and A.~D. Dragan, ``Sirl: Similarity-based implicit representation learning,'' in \emph{Proceedings of the 2023 ACM/IEEE International Conference on Human-Robot Interaction}, 2023, pp. 565--574.

\bibitem{rafailov2024direct}
R.~Rafailov, A.~Sharma, E.~Mitchell, C.~D. Manning, S.~Ermon, and C.~Finn, ``Direct preference optimization: Your language model is secretly a reward model,'' \emph{Advances in Neural Information Processing Systems}, vol.~36, 2024.

\bibitem{myers2022learning}
V.~Myers, E.~Biyik, N.~Anari, and D.~Sadigh, ``Learning multimodal rewards from rankings,'' in \emph{Conference on Robot Learning}.\hskip 1em plus 0.5em minus 0.4em\relax PMLR, 2022, pp. 342--352.

\bibitem{shin2023benchmarks}
D.~Shin, A.~D. Dragan, and D.~S. Brown, ``Benchmarks and algorithms for offline preference-based reward learning,'' \emph{arXiv preprint arXiv:2301.01392}, 2023.

\bibitem{liu2023efficient}
Y.~Liu, G.~Datta, E.~Novoseller, and D.~S. Brown, ``Efficient preference-based reinforcement learning using learned dynamics models,'' \emph{International Conference on Robotics and Automation (ICRA)}, 2023.

\bibitem{wilde2021learning}
N.~Wilde, E.~B{\i}y{\i}k, D.~Sadigh, and S.~L. Smith, ``Learning reward functions from scale feedback,'' \emph{arXiv preprint arXiv:2110.00284}, 2021.

\bibitem{richards2005mixed}
A.~Richards and J.~How, ``Mixed-integer programming for control,'' in \emph{Proceedings of the 2005, American Control Conference, 2005.}\hskip 1em plus 0.5em minus 0.4em\relax IEEE, 2005, pp. 2676--2683.

\bibitem{belotti2016handling}
P.~Belotti, P.~Bonami, M.~Fischetti, A.~Lodi, M.~Monaci, A.~Nogales-G{\'o}mez, and D.~Salvagnin, ``On handling indicator constraints in mixed integer programming,'' \emph{Computational Optimization and Applications}, vol.~65, pp. 545--566, 2016.

\bibitem{montazery2017dominance}
M.~Montazery and N.~Wilson, ``Dominance and optimisation based on scale-invariant maximum margin preference learning.''\hskip 1em plus 0.5em minus 0.4em\relax International Joint Conferences on Artificial Intelligence, 2017.

\bibitem{teso2016constructive}
S.~Teso, A.~Passerini, and P.~Viappiani, ``Constructive preference elicitation by setwise max-margin learning,'' \emph{arXiv preprint arXiv:1604.06020}, 2016.

\bibitem{yuan2015non}
X.~Yuan, R.~Henao, E.~Tsalik, R.~Langley, and L.~Carin, ``Non-gaussian discriminative factor models via the max-margin rank-likelihood,'' in \emph{International Conference on Machine Learning}.\hskip 1em plus 0.5em minus 0.4em\relax PMLR, 2015, pp. 1254--1263.

\bibitem{gymnasium_robotics2023github}
\BIBentryALTinterwordspacing
R.~de~Lazcano, K.~Andreas, J.~J. Tai, S.~R. Lee, and J.~Terry, ``Gymnasium robotics,'' 2023. [Online]. Available: \url{http://github.com/Farama-Foundation/Gymnasium-Robotics}
\BIBentrySTDinterwordspacing

\bibitem{haarnoja2018soft}
T.~Haarnoja, A.~Zhou, K.~Hartikainen, G.~Tucker, S.~Ha, J.~Tan, V.~Kumar, H.~Zhu, A.~Gupta, P.~Abbeel, \emph{et~al.}, ``Soft actor-critic algorithms and applications,'' \emph{arXiv preprint arXiv:1812.05905}, 2018.

\bibitem{todorov2012mujoco}
E.~Todorov, T.~Erez, and Y.~Tassa, ``Mujoco: A physics engine for model-based control,'' in \emph{2012 IEEE/RSJ international conference on intelligent robots and systems}.\hskip 1em plus 0.5em minus 0.4em\relax IEEE, 2012, pp. 5026--5033.

\bibitem{chen2020simple}
T.~Chen, S.~Kornblith, M.~Norouzi, and G.~Hinton, ``A simple framework for contrastive learning of visual representations,'' in \emph{International conference on machine learning}.\hskip 1em plus 0.5em minus 0.4em\relax PMLR, 2020, pp. 1597--1607.

\bibitem{mattson2023leveraging}
C.~Mattson and D.~S. Brown, ``Leveraging human feedback to evolve and discover novel emergent behaviors in robot swarms,'' \emph{Genetic and Evolutionary Computation Conference (GECCO)}, 2023.

\end{thebibliography}

\appendices
\onecolumn

\section{PBICRL: Unknown features}
\setcounter{algorithm}{1}

\subsection{Constraint Inference with Unknown Feature Parameters}
This section contains the algorithm for the variation of PBICRL that further allows for inferring the parametric form of the constraints. More specifically, we assume that the $M$ constraint features $\boldsymbol{\phi}^j_{{\theta}},\;j=1,\ldots,M,$ have now functional forms parameterized by ${\theta}$. In this case, we assume that the environment reward is given by
\begin{align}
r_{\boldsymbol{\theta}}(s)=r_{n}(s)+r_{p,\theta}(s)=\mathbf{w}_{n}^{\top}\boldsymbol{\phi}_n(s) + (\mathbf{c}_p\circ\mathbf{w}_{p})^{\top}\boldsymbol{\phi}_{{\theta}}(s),
\end{align}
with $\mathbf{w}_{p}\in\mathbb{R}^M$, $\mathbf{c}_p\in\mathbb{R}^M$, being unknown and $\mathbf{w}_n\in\mathbb{R}^{N_{\phi}}$ still corresponding to the known nominal weight vector of the known features, which are now denoted with $\boldsymbol{\phi}_n$. Algorithm~\ref{alg:feat_inf_unknown} demonstrates the new sampling scheme, in which the also unknown constraint parameters ${\theta}\in\mathbb{R}^{N_{\theta}}$ are sampled and included in the calculation of the likelihood function
\begin{align}\label{eq:likeli_fn_unknwon_feat}
    \mathcal{L}(\mathbf{w}_{p}, \mathbf{c}_{p},\boldsymbol{\theta})=&\hspace{-0.5cm}\sum\limits_{\substack{\forall \tau_i\in\mathcal{G}_k,\forall \tau_j\in\mathcal{G}_{\ell}\\\forall k<\ell}} \hspace{-0.5cm}log P(\tau_i\succ\tau_j)\nonumber \\=&\hspace{-0.5cm}\sum\limits_{\substack{\forall \tau_i\in\mathcal{G}_k,\forall \tau_j\in\mathcal{G}_{\ell}\\\forall k<\ell}} \hspace{-0.5cm}log\frac{e^{\frac{\beta}{T_i}\sum_{s\in\tau_i }{r}_{\boldsymbol{\theta}}(s)-m_{k\ell}}}{e^{\frac{\beta}{T_i}\sum_{s\in\tau_i}{r}_{\boldsymbol{\theta}}(s)-m_{k\ell}}+e^{\frac{\beta}{T_j}\sum_{s\in\tau_j}{r}_{\boldsymbol{\theta}}(s)}}.
\end{align}
In the above formulations, we use $\boldsymbol{\theta}$  to denote all the unknown parameters and $\theta$ to denote just the unknown feature parameters.
\begin{algorithm}[H]
\caption{PBICRL (Parametric Features)}\label{alg:feat_inf_unknown}
\begin{algorithmic}[1]
\State \textbf{Parameters:} Number of iterations $k$, sampling frequency $f_s$, $\sigma_1$, $\sigma_2$, margins $m_{k\ell}, \forall k\neq\ell\in \{1,\ldots, K\}$
\State Randomly sample: penalty weight $\mathbf{w}_{p}\in\mathbb{R}^M$, constraint feature $\mathbf{c}_{p}\in\{0,1\}^M$, $\theta\in\mathbb{R}^{N_{\theta}}$
\For{$i=1,\ldots,k$ }
\State Randomly sample feature $j$ from $\{1,\ldots,M\}$
\If{{$i$ mod $f_s$!=0}}
\State Set $\mathbf{c}_{p}'[j]=\lnot \mathbf{c}_{p}[j]$, $\mathbf{w}_{p}'=\mathbf{w}_{p}$, ${\theta}'= {\theta}$
\ElsIf{{$i$ mod $f_s$!=1}}
\State Set $\mathbf{w}_{p}'[j]= \mathbf{w}_{p}[j]+\mathcal{N}(0,\sigma_1)$, $\mathbf{c}_{p}'=\mathbf{c}_{p}$ , ${\theta}'= {\theta}$
\Else
\State Set ${\theta}_{j}'= {\theta}_{j}+\mathcal{N}(\mathbf{0},\sigma_2\mathbb{I}_{N_{\theta}})$, $\mathbf{c}_{p}'=\mathbf{c}_{p}$ , $\mathbf{w}_{p}'=\mathbf{w}_{p}$
\EndIf
\State Compute Likelihood using~(\ref{eq:likeli_fn_unknwon_feat})
\If{$\log\mathcal{L}(\mathbf{w}_{p}',\mathbf{c}_{p}',{\theta}')\geq \log\mathcal{L}(\mathbf{w}_{p},\mathbf{c}_{p},{\theta})$}
\State Set $\mathbf{w}_{p}=\mathbf{w}_{p}'$, $\mathbf{c}_{p}=\mathbf{c}_{p}'$, ${\theta}={\theta}'$ 
\Else
\State Set $\mathbf{w}_{p}=\mathbf{w}_{p}'$, $\mathbf{c}_{p}=\mathbf{c}_{p}'$, ${\theta}={\theta}'$ w.p. $\mathcal{L}(\mathbf{w}_{p}',\mathbf{c}_{p}',{\theta}')/\mathcal{L}(\mathbf{w}_{p},\mathbf{c}_{p},{\theta})$
\EndIf
\EndFor
\State \textbf{Return} ${\mathbf{w}_{p}}, \mathbf{c}_{p}$, ${\theta}$
\end{algorithmic}
  \phantomcaption

\end{algorithm}

\section{Experiments}\label{sec:sims_app}
We utilize four different environments to quantify the performance of PBICRL: \textit{(1)} a 2D point mass navigation environment, \textit{(2)} the Fetch-Reach environment from OpenAI Gym and \textit{(3)} the HalfCheetah and \textit{(4)} Ant environments provided in Gymnasium~\cite{gymnasium_robotics2023github}. Demonstrations in all environments were obtained from policies obtained with the Soft Actor-Critic (SAC) algorithm~\cite{haarnoja2018soft}.
\begin{figure}[H]
\centering
        \begin{subfigure}[t]{0.25\textwidth}
                \centering   \includegraphics[width=1.21\textwidth]{figures/point_mass_env.png}
                \vspace{-0.7cm}
                \caption{}    \label{fig:2d_env_2}
        \end{subfigure}%
        \begin{subfigure}[t]{0.25\textwidth}
        \vspace{-3.7cm}
                \centering    \includegraphics[width=0.82\textwidth]{figures/fetch.pdf}
                \caption{}    \label{fig:fetch_env_2}
        \end{subfigure}%
        \begin{subfigure}[t]{0.25\textwidth}
        \vspace{-3.7cm}
                \centering    \includegraphics[width=0.8\textwidth]{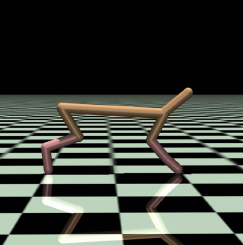}
                \caption{}     \label{fig:cheetah_env}
                
        \end{subfigure}%
        \begin{subfigure}[t]{0.25\textwidth}
        \vspace{-3.7cm}
                \centering     \includegraphics[width=0.8\textwidth]{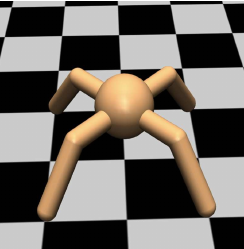}
                \caption{}     \label{fig:ant_env}
        \end{subfigure}%
         \caption{2D point mass navigational environment (\subref{fig:2d_env_2}), FecthReach robot (\subref{fig:fetch_env_2}), HalfCheetah (\subref{fig:cheetah_env}) and Ant (\subref{fig:ant_env}) simulation environments.}
\label{fig:point_mass_comp_100}
\end{figure}

\subsection{Point Mass and Fetch-Reach Robot}

\begin{figure}[H]
\centering
\begin{subfigure}[t]{0.33\textwidth}
                \centering   
                \includegraphics[width=0.89\textwidth]{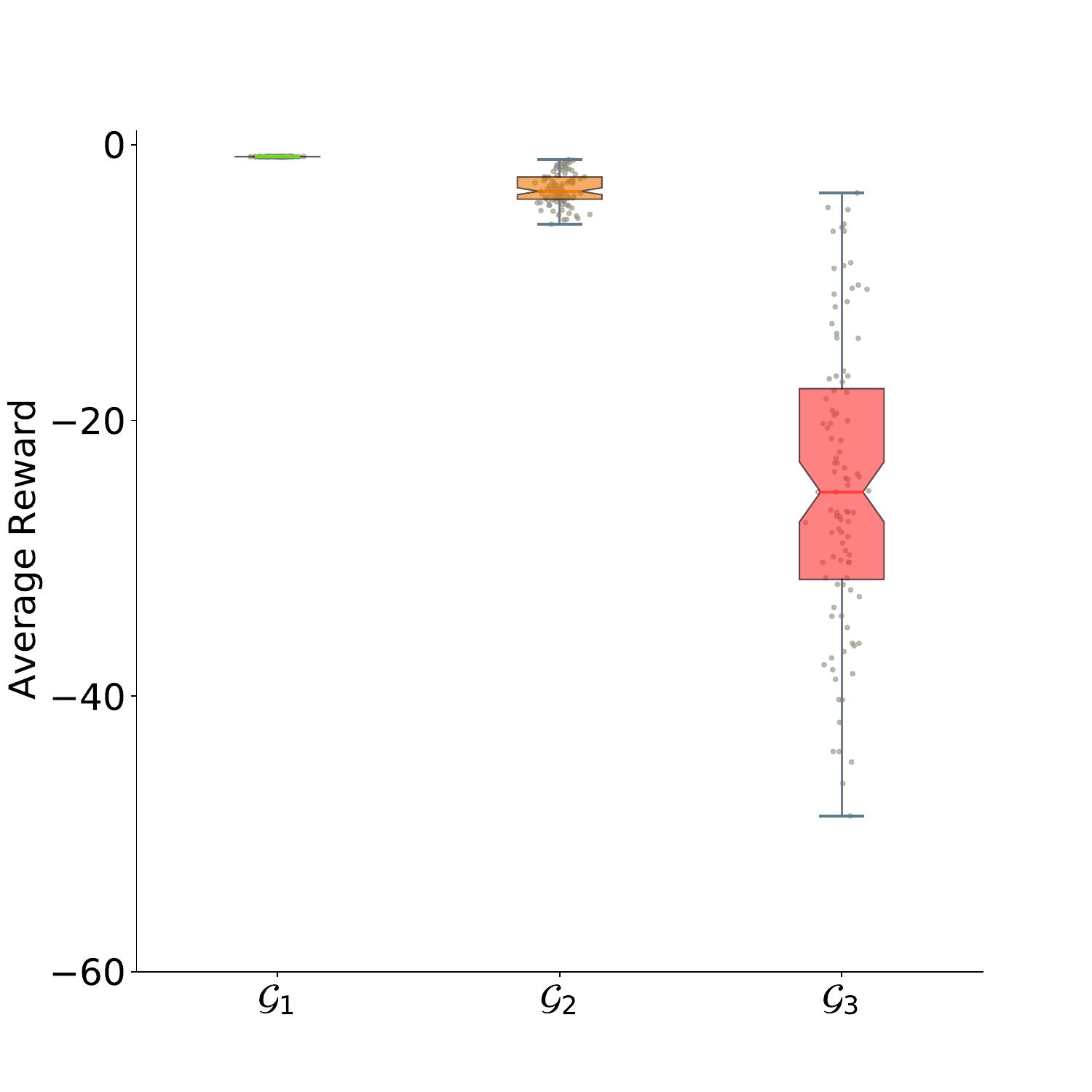}
                \caption{}                \label{fig:marg_res1}
        \end{subfigure}%
        \begin{subfigure}[t]{0.33\textwidth}
                \centering   
                \includegraphics[width=0.89\textwidth]{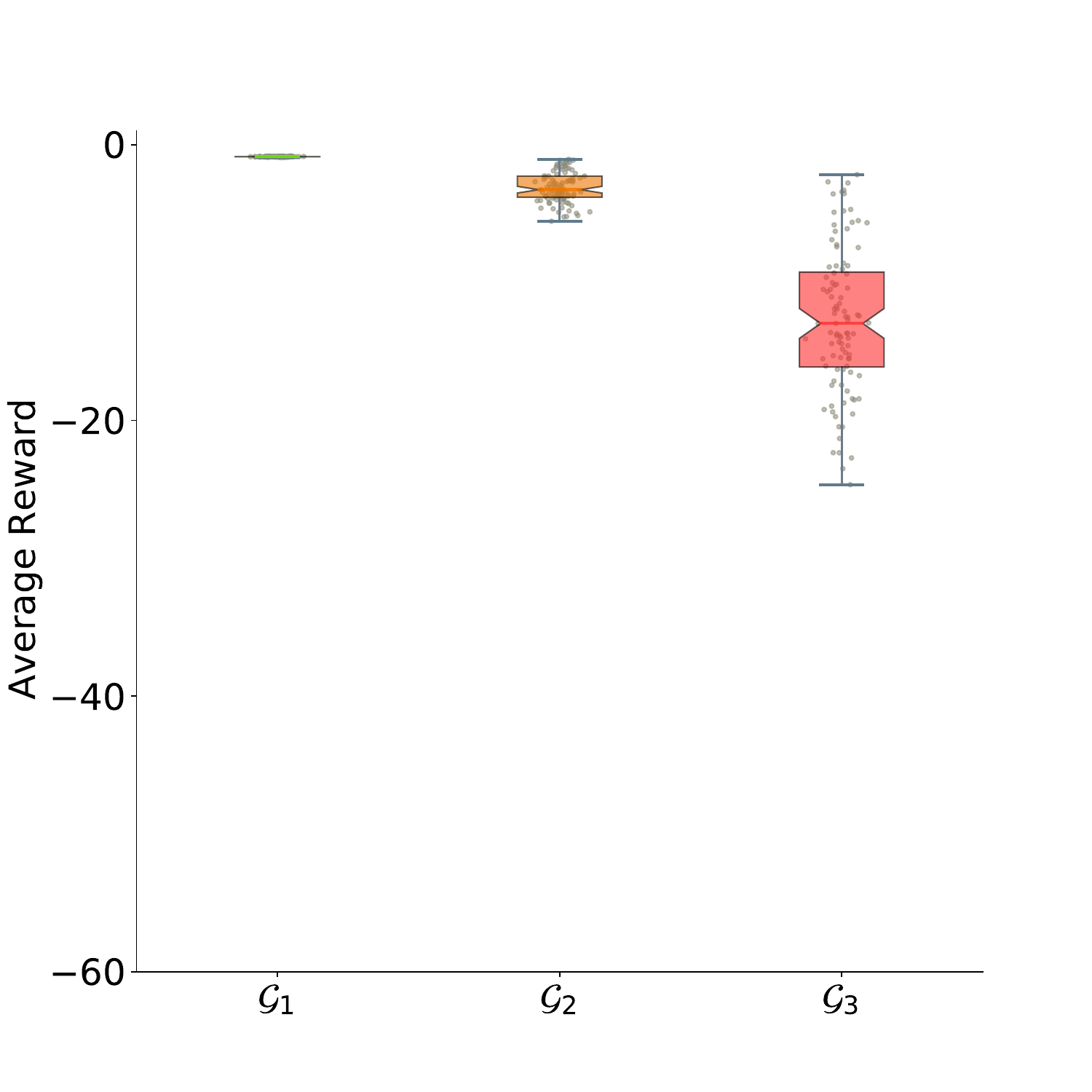}
                \caption{}
                \label{fig:marg_res2}
         \end{subfigure}%
         \begin{subfigure}[t]{0.33\textwidth}
                \centering   
                \includegraphics[width=0.89\textwidth]{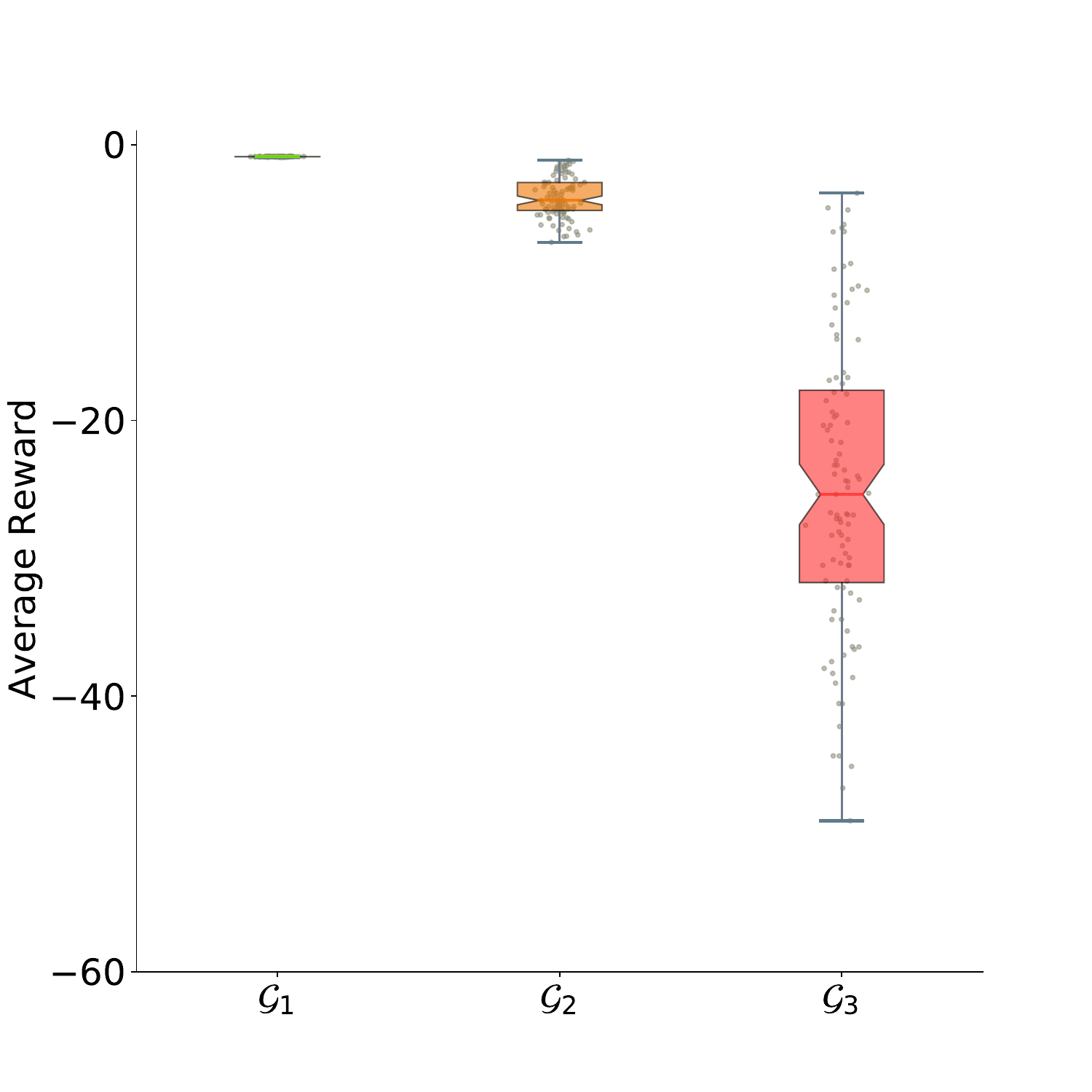}
                \caption{}
                \label{fig:marg_res3}
         \end{subfigure}%
         \caption{Reward distribution of $60$ trajectories in each of the three demonstration groups under the true weights~(\subref{fig:marg_res1}), PBICRL without margins~(\subref{fig:marg_res2}) and PBICRL with margins~(\subref{fig:marg_res3}). Lateral noise is added on demonstration rewards (grey points) for better visualization.}
\label{fig:cont_human_priortt}
\end{figure}

This section contains details on tuning the margins $m_{ij}$ for the point mass and Fetch-Reach environments. To tune these margins we assume access to the information that  the demonstrations in $\mathcal{G}_2$ are approximately four times more preferable to those in $\mathcal{G}_3$ than those in $\mathcal{G}_1$ are to those in $\mathcal{G}_2$. This can be seen in Figure~\ref{fig:marg_res1}, which shows the mean of the distribution of the rewards of the demonstrations in each group when both nominal and penalty weights are known. The difference between the mean values for $\mathcal{G}_2$ and $\mathcal{G}_3$ is approximately four times that between $\mathcal{G}_1$ and $\mathcal{G}_2$. Figure~\ref{fig:marg_res2} shows the distribution of rewards per group based on the inferred $\mathbf{w}_p$ and $\mathbf{c}_p$ from PBICRL with no margins ($m_{ij}=0,\forall i,j$). It is clear that the original margins are not respected by simply using the Bradley-Terry preference model. 

Having as target to satisfy the given relationship between the margins we tune the $m_{ij}$ parameters to $m_{12}=0.0$, $m_{23}=2.9$ and $m_{13}=6.0$ in order to match those margins. Figure~\ref{fig:marg_res3} shows the distribution of the cumulative rewards of the given demonstrations under the inferred penalty weights. Tuning was done by first selecting a set of candidate values for the margin parameters $m_{ij}$. Then for each of those sets, the constraints were inferred and the margins between the mean values of the distributions of the rewards for the demonstrations in each group we calculated. The $m_{ij}$ values that resulted in margins close to the real ones were kept. 

\begin{figure}[H]
\centering
\begin{subfigure}[t]{0.33\textwidth}
                \centering   
                \includegraphics[width=0.89\textwidth]{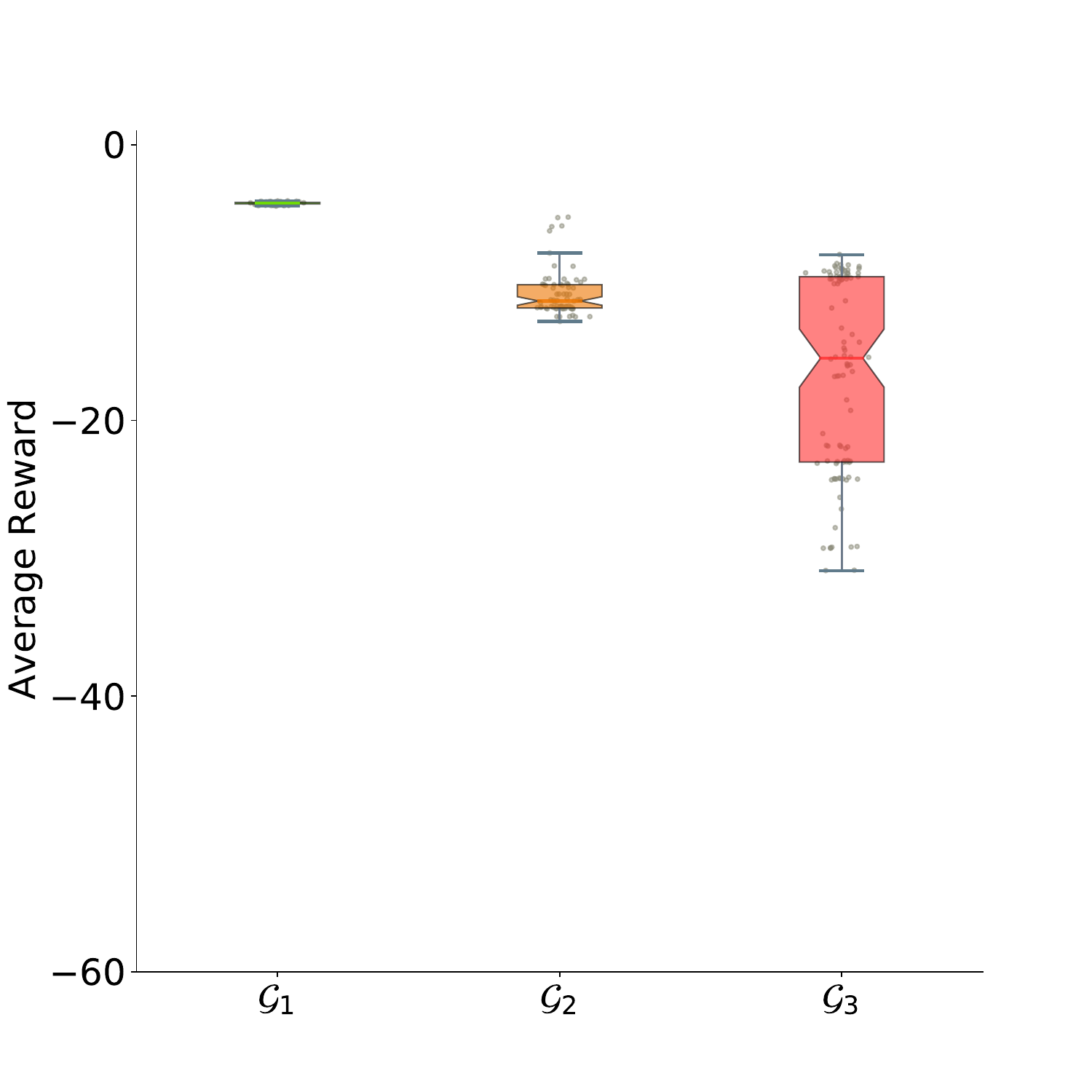}
                \caption{}                \label{fig:fetch_marg_res1}
        \end{subfigure}%
        \begin{subfigure}[t]{0.33\textwidth}
                \centering   
                \includegraphics[width=0.89\textwidth]{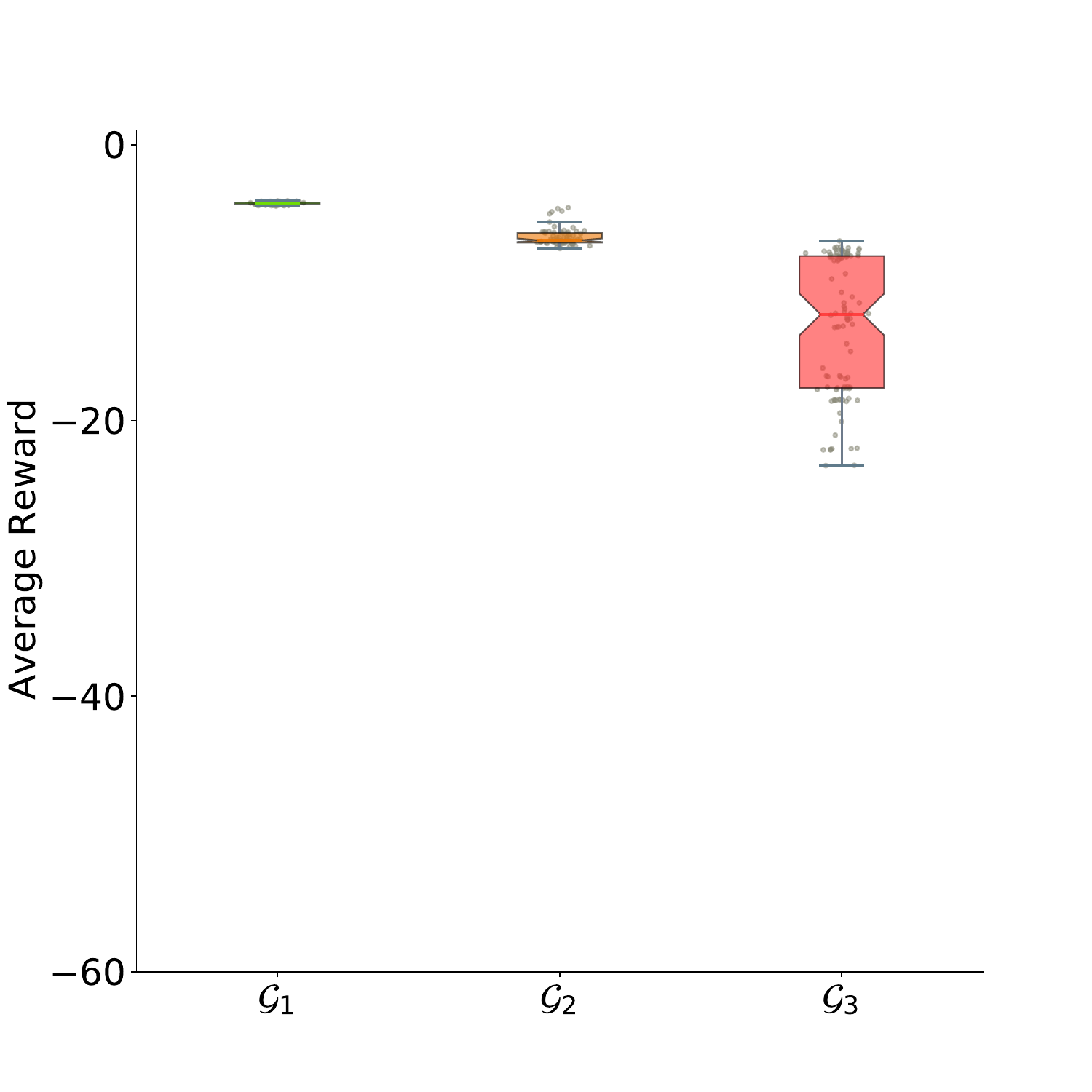}
                \caption{}
                \label{fig:fetch_marg_res2}
         \end{subfigure}%
         \begin{subfigure}[t]{0.33\textwidth}
                \centering   
                \includegraphics[width=0.89\textwidth]{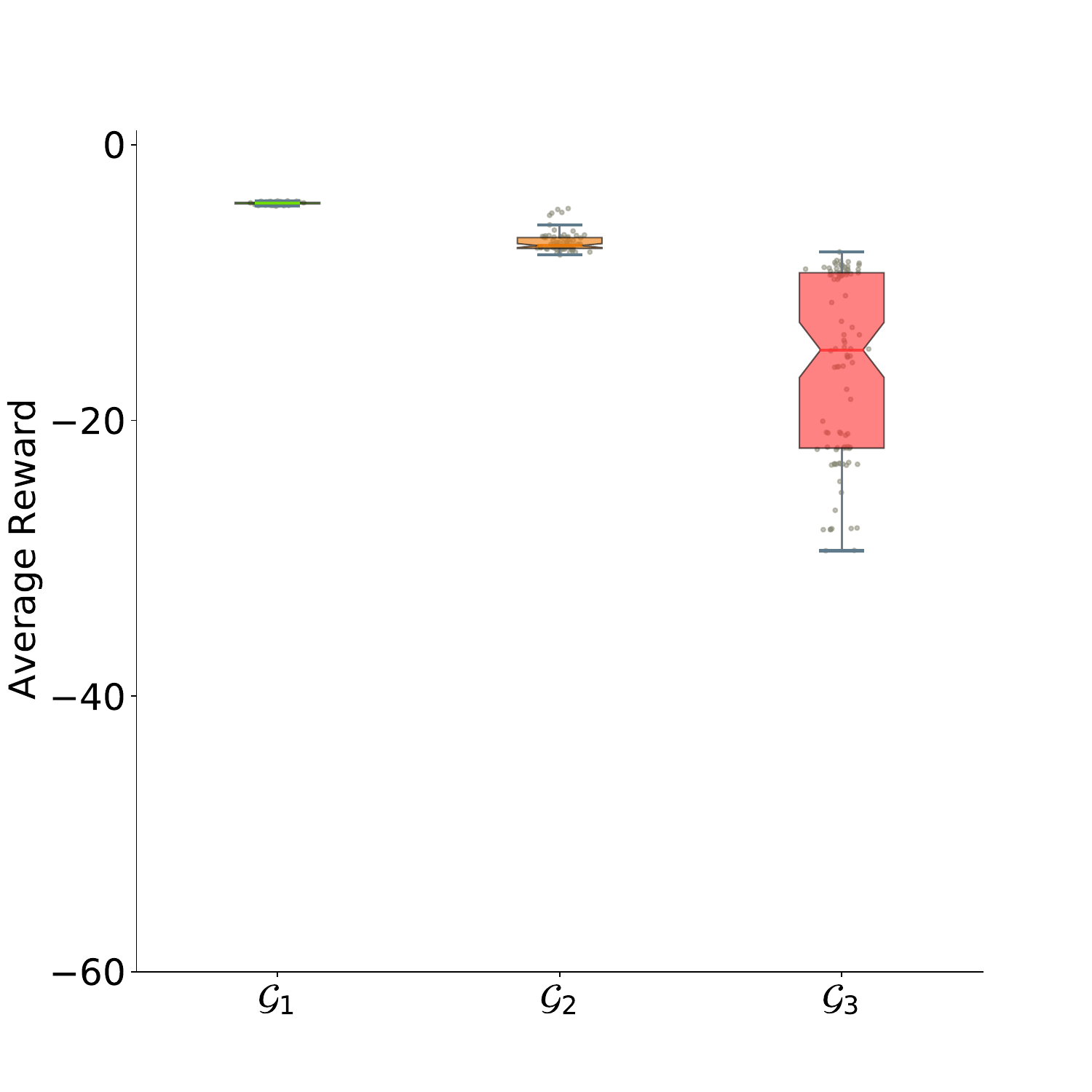}
                \caption{}
                \label{fig:fetch_marg_res3}
         \end{subfigure}%
         \caption{Reward distribution of $60$ trajectories in each of the three demonstration groups under the true weights~(\subref{fig:fetch_marg_res1}), PBICRL with no margins (\subref{fig:fetch_marg_res2}) and PBICRL with margins~(\subref{fig:fetch_marg_res3}).}
\label{fig:fetch_marg_plots}
\end{figure}

Similarly, in the Fetch-Reach environment the $m_{ij}$ parameters were tuned under the assumption that the margins between the two pairs of groups $\mathcal{G}_1, \mathcal{G}_2$ and $\mathcal{G}_2, \mathcal{G}_3$ are approximately the same. The distribution of the original rewards of the demonstrations, along with the ones for the inferred weights under $m_{ij}=0, \forall i,j$ and after tuning $m_{ij}$ to $m_{12}=1.0$, $m_{23}=1.5$ and $m_{13}=2.5$, can be seen in Figure~\ref{fig:fetch_marg_plots}.

\subsection{HalfCheetah and Ant with Parametric Constraints}
\begin{figure}[H]
\centering
\begin{subfigure}[t]{0.5\textwidth}
                \centering   
                \includegraphics[scale=0.12]{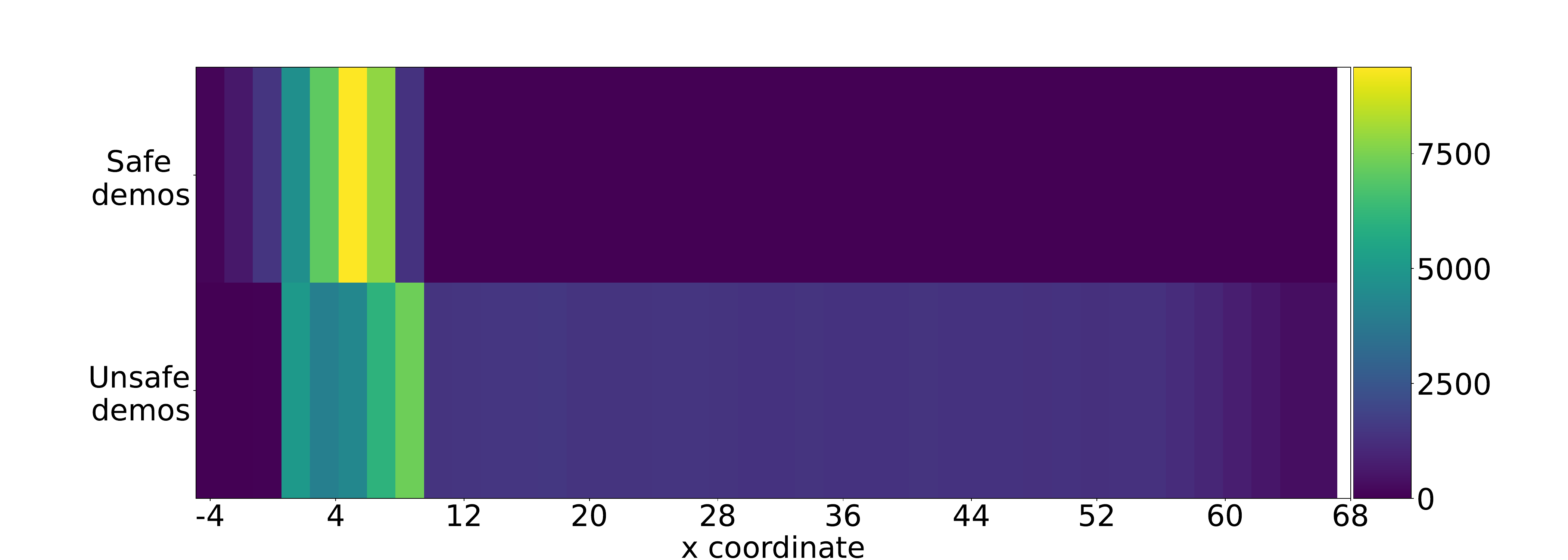}
                \caption{}                \label{fig:cheetah_traj}
        \end{subfigure}%
        \begin{subfigure}[t]{0.5\textwidth}
                \centering   
                \includegraphics[scale=0.12]{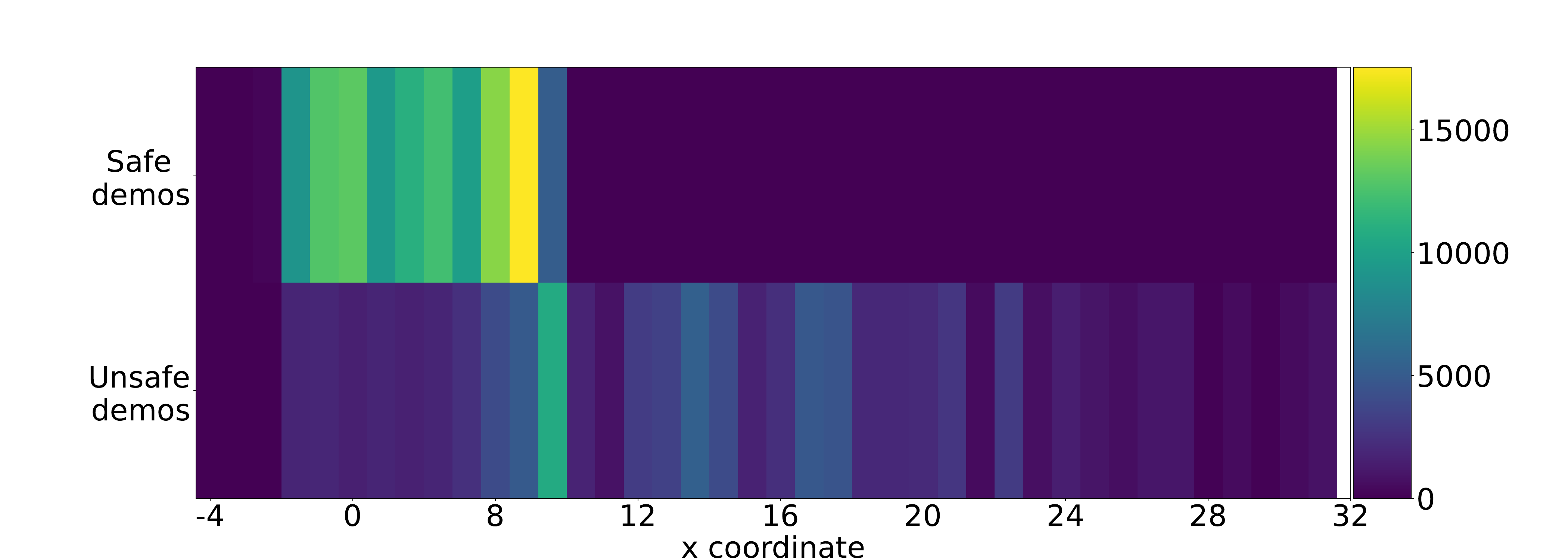}
                \caption{}
                \label{fig:ant_traj}
         \end{subfigure}%
         \caption{Visitation frequencies of good and bad demonstrations for HalfCheetah~(\subref{fig:cheetah_traj}) and Ant~(\subref{fig:ant_traj}) environments. The domain has been discretized just for exposition purposes.}
\label{fig:cheetah_ant_traj}
\end{figure}

In this section, we consider parametric 1D halfspace constraint functions of the form $z\leq \theta_i$, where PBICRL must also infer the parameter $\theta_i\in\mathbb{R}^1$.
We assume that there exists a constraint inhibiting the agent's forward motion at a particular location $z$.  For the HalfCheetah we assume that the agent faces an obstacle at the $z=8$ location and whenever $z\geq 8$ the agent incurs a penalty reward of $-50$. In the case of the Ant environment, we assume that the states for which $z\geq 10$ are constrained and associated with a penalty reward of $-100$. Figure~\ref{fig:cheetah_ant_traj} depicts the visitation frequencies of $100$ good and $100$ bad, or safe and constraint violating, trajectories of length $500$ each for the two environments.   

\section{Simulation Details}

\subsection{Soft Actor-Critic Algorithm}\label{sec:app_soft_actor_critic}
The policies in the environment were obtained using a SAC policy~\cite{haarnoja2018soft}. The hyperparameters of the policy training can be seen in the Table~\ref{table:SAC_params}. For the HalfCheetah and Ant environments the maximum episode length was $1000$. In the first $40$ episodes actions were chosen randomly to enhance exploration. The three SAC learning rates are denoted with $\lambda_V, \lambda_Q,\lambda_{\pi}$.
\begin{table}[H]
\caption{Hyperparameters of SAC training for Point Mass (PM), Fetch-Reach (FR), HalfCheetah (HC) and Ant environments.}
\centering
\resizebox{0.45\textwidth}{!}{
\begin{tabular}{cccccc}
\thickhline
 \multicolumn{3}{c}{Hyperparameter} & \multicolumn{3}{c}{Value}\\ \cmidrule(lr){1-3} \cmidrule(lr){4-6}
   & $\lambda_V$ &  &  & $3\cdot 10^{-4}$ &  \\ 
 & $\lambda_Q$ &  &  & $3\cdot 10^{-4}$ &  \\ 
   & $\lambda_{\pi}$ &  &  & $3\cdot 10^{-4}$ &  \\ 
  & batch size &  &  & $128$ &  \\ 
    & hidden dimension &  &  & $256$ &  \\ 
     & hidden layers &  &  & $2$ &  \\ 
      & buffer size &  &  & $10000$ &  \\ 
       & max steps &  &  & $500$ &  \\
      & Iterations (PM)&  &  & $200000$ & \\
      & Iterations (FR)&  &  & $250000$ &\\
      & Iterations (HC, Ant)&  &  & $500000$ &\\
 \thickhline
\end{tabular}}
\label{table:SAC_params}
\end{table}

\subsection{BPL}\label{sec:birl_app}
The following MCMC algorithm is the baseline feature-based BPL algorithm~\cite{brown2020safe} used in the experiments for the point mass and Fetch-Reach environment comparisons, as well as the illustration example. In the case of BPL, the reward function is given by
\begin{align}
r_{\boldsymbol{\theta}}(s)=r_{n}(s)+r_{p}(s)=(\mathbf{w}_{n}+\mathbf{w}_{p})^{\top}\boldsymbol{\phi}(s),
\end{align}
with the main difference with the reward function assumed in PBICRL being the omission of the binary variables $\mathbf{c}_p$. For completeness, the likelihood function used in Algorithm~\ref{alg:birl} is
\begin{align}\label{eq:likeli_fn_app}
    \mathcal{L}(\mathbf{w}_{p},\mathbf{c}_p)&=\sum_{\tau_i\succ\tau_j} \log P(\tau_i\succ\tau_j)\nonumber\\&=\sum_{\tau_i\succ\tau_j}\log\frac{e^{\frac{\beta}{T_i}\sum_{s\in\tau_i}r_{\boldsymbol{\theta}}(s)}}{e^{\frac{\beta}{T_i}\sum_{s\in\tau_i}r_{\boldsymbol{\theta}}(s)}+e^{\frac{\beta}{T_j}\sum_{s\in\tau_j}r_{\boldsymbol{\theta}}(s)}}.
\end{align}
\setcounter{algorithm}{2}

\begin{algorithm}[H]
\caption{BPL}\label{alg:birl}
\begin{algorithmic}[1]
\State \textbf{Parameters:} Number of iterations $k$, $\sigma$
\State Randomly sample: penalty weight $\mathbf{w}_{p}\in\mathbb{R}^{N_{\phi}}$
\For{$i=1,\ldots,k$ }
\State Randomly sample feature $j$ from $\{1,\ldots,N_{\phi}\}$
\State Set $\mathbf{w}_{p}'[j]= \mathbf{w}_{p}[j]+\mathcal{N}(0,\sigma)$
\State Compute Likelihood using~(\ref{eq:likeli_fn_app})
\If{$\log\mathcal{L}(\mathbf{w}_{p}')\geq \log\mathcal{L}(\mathbf{w}_{p})$}
\State Set $\mathbf{w}_{p}=\mathbf{w}_{p}'$
\Else
\State Set $\mathbf{w}_{p}=\mathbf{w}_{p}'$ w.p. $\mathcal{L}(\mathbf{w}_{p}')/\mathcal{L}(\mathbf{w}_{p})$
\EndIf
\EndFor
\State \textbf{Return} $\mathbf{w}_{p}$
\end{algorithmic}
\end{algorithm}

\subsection{PBICRL Hyperparameters}\label{sec:pbirl_app}
The hyperparameters used in the results regarding PBICRL can be seen in Table~\ref{table:pbicrl_params}. Initialization of all parameters was done randomly.
\begin{table}[H]
\caption{Hyperparameters of PBICRL used in Point Mass (PM), Fetch-Reach (FR), HalfCheetah (HC) and Ant environments simulations.}
\centering
\resizebox{0.4\textwidth}{!}{
\begin{tabular}{cccccc}
\thickhline
 \multicolumn{3}{c}{Hyperparameter} & \multicolumn{3}{c}{Value}\\ \cmidrule(lr){1-3} \cmidrule(lr){4-6}
 & $k$ &  &  & $4\cdot 10^5$ &  \\ 
   & $\sigma$ (PM, FR) &  &  & $0.1$ &  \\ 
    & $f_s$ (PM, FR)&  &  & $4$ &  \\ 
    & $f_s$ (HC, Ant)&  &  & $3$ &  \\ 
 & $\sigma_1$ (HC, Ant) &  &  & $1.0$ &  \\ 
 & $\sigma_2$ (HC, Ant) &  &  & $0.5$ &  \\ 
 \thickhline
\end{tabular}}
\label{table:pbicrl_params}
\end{table}

\section{Additional Results}
This section contains additional simulation results that quantify the performance of PBICRL. We carry out sensitivity analysis with respect of the number of demonstrations provided for each group $\mathcal{G}_i$. We present results only for PBICRL with $m_{ij}=0$.

\subsection{Point Mass and Fetch-Reach Environments}

We vary the number of demonstrations per group $\mathcal{G}_i, i=1,2,3$, to $20$, $60$ and $100$ in order to quantify the effect of the number of demonstrations on the inferred weights. As expected, the performance of PBICRL improves with the number of demonstrations.
\begin{figure}[H]
\centering
\begin{subfigure}[t]{0.5\textwidth}
                \centering   \includegraphics[width=0.79\textwidth]{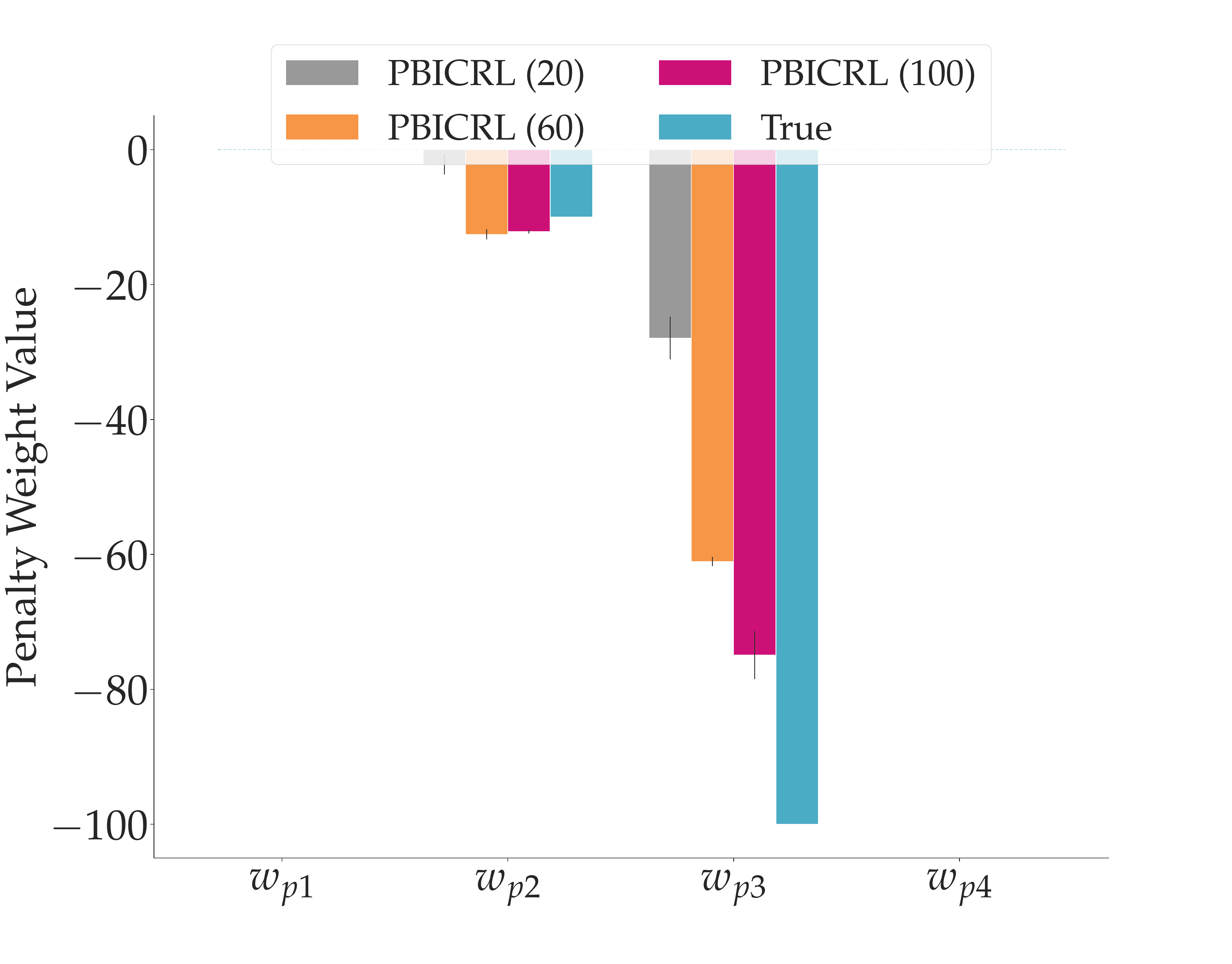}
                \caption{}    \label{fig:point_mass_multiple}
        \end{subfigure}%
        \begin{subfigure}[t]{0.5\textwidth}
                \centering     \includegraphics[width=0.79\textwidth]{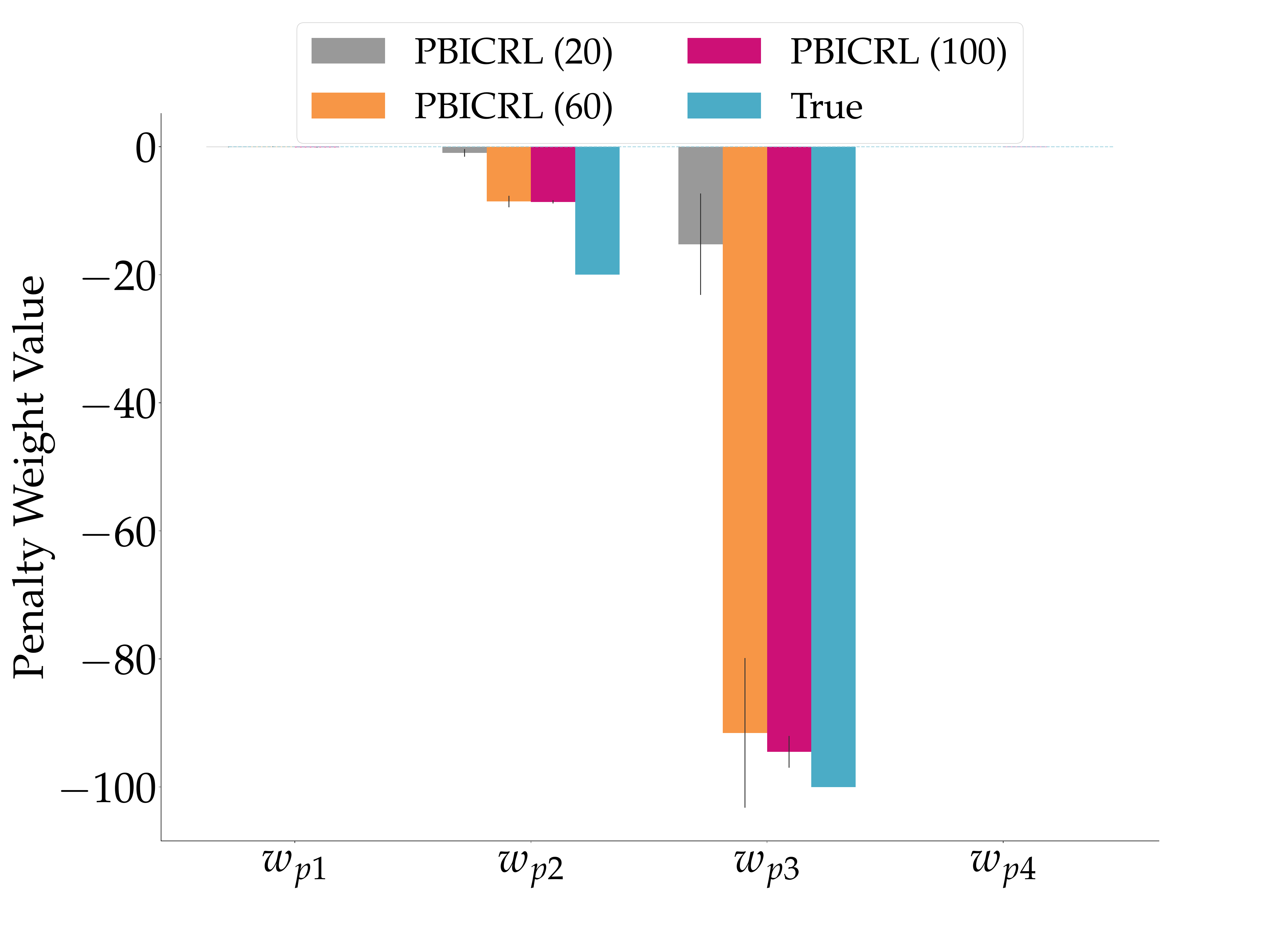}
                \caption{}     \label{fig:fetch_reach_multiple}
         \end{subfigure}%
         \caption{Inferred weights for $20$, $60$ and $100$ demonstrations per group for point mass (\subref{fig:point_mass_multiple}) and Fetch-Reach (\subref{fig:fetch_reach_multiple}) environments, respectively. Results are averaged over $5$ seeds.}
\label{fig:point_mass_fetch_multiple}
\end{figure}

\subsection{HalfCheetah and Ant Environments}
We similarly vary the number of demonstrations provided in the two groups $\mathcal{G}_1$ and $\mathcal{G}_2$ for the HalfCheetah and Ant environments and report the mean and standard deviation of  CMSE in Table~\ref{table:cmse_multiple}. Clearly a larger number of demonstrations leads to better inference.
\begin{table}[H]
\caption{CMSE for PBICRL with varying number of demonstrations. Results are averaged over $5$ seeds.}
\centering
\resizebox{0.5\textwidth}{!}{
\begin{tabular}{cccccccc}
\thickhline
 \# demos&& & Env. & & \multicolumn{3}{c}{PBICRL}\\  \cmidrule(lr){1-2}  \cmidrule(lr){3-5} \cmidrule(lr){6-8} 
 $20$ & & &HC & &   & ${50.68}\pm{98.68}$  &  \\ 
 && &Ant &   &  & {$0.78$} $\pm$ {$0.54$}  &  \\ 
 \hline
  $50$ &&  &HC & &   & ${0.13}\pm{0.17}$  &  \\ 
 &&& Ant &   &  & {$0.41$} $\pm$ {$0.41$}  &  \\ 
 \hline
  $100$ &&  &HC & &   & ${0.073}\pm{0.12}$  &  \\ 
 &&& Ant &   &  & {$0.20$} $\pm$ {$0.29$}  &  \\ 
 \hline
 $200$ &&  &HC & &   & ${0.017}\pm{0.017}$  &  \\ 
 &&& Ant &   &  & {$0.011$} $\pm$ {$0.009$}  &  \\ 
 \thickhline
\end{tabular}}
\label{table:cmse_multiple}
\end{table}


\end{document}